\documentclass[11pt]{article}
\PassOptionsToPackage{table}{xcolor}
\usepackage{fontawesome5}
\PassOptionsToPackage{hyperfootnotes=false}{hyperref}
\usepackage[preprint]{acl}
\usepackage{tabularx}
\usepackage{times}
\usepackage{latexsym}
\usepackage[T1]{fontenc}
\usepackage[utf8]{inputenc}
\usepackage[expansion=false]{microtype}
\usepackage{inconsolata}
\usepackage{graphicx}
\usepackage{amsmath}
\usepackage{amssymb}
\usepackage{booktabs}
\usepackage{multirow}
\usepackage{array}
\usepackage{xcolor}
\usepackage{bm}
\usepackage{pifont}
\usepackage{placeins}
\usepackage{float}
\usepackage{tikz}
\usepackage[most]{tcolorbox}

\definecolor{tablehead}{HTML}{F4EFFF}
\definecolor{tablealt}{HTML}{FAF8FE}
\definecolor{oursfill}{HTML}{EAF3FF}
\definecolor{promptline}{HTML}{8877A3}
\definecolor{promptfill}{HTML}{FBF9FE}
\definecolor{prompthead}{HTML}{EDE2FF}
\definecolor{todofill}{HTML}{FAECCA}
\definecolor{todoline}{HTML}{BBA262}
\definecolor{todoink}{HTML}{534624}

\newtcolorbox{promptbox}[1]{enhanced, before upper=\raggedright, colback=promptfill, colframe=promptline,
  boxrule=0.4pt, arc=1.5pt, left=4pt, right=4pt, top=2.5pt, bottom=2.5pt,
  colbacktitle=prompthead, coltitle=black, fonttitle=\small\bfseries, title={#1},
  fontupper=\footnotesize, before skip=6pt, after skip=6pt}

\newif\ifexptodo \exptodofalse

\newif\ifdraft \draftfalse
\ifdraft
  
  \newcommand{\tbd}[1]{{\color{blue!60!black}\textbf{[TBD}\if\relax\detokenize{#1}\relax\else{: #1}\fi\textbf{]}}}
\else
  
  \newcommand{\tbd}[1]{\errmessage{Unresolved TBD: #1}}
\fi
\ifdraft\else\fi
\newcommand{\na}{\textcolor{black!45}{--}}

\newcommand{\gain}[1]{\textsubscript{\,\textcolor{green!45!black}{#1}}}
\newcommand{\dmid}{Duopra-RL1}
\newcommand{\tool}[1]{\texttt{#1}}
\newcommand{\tok}[1]{\texttt{#1}}
\newcommand{\Fcat}{\mathcal{F}}
\newcommand{\FB}{\mathcal{F}_{B}}
\newcommand{\Cgold}{C^{\ast}}

\newcommand{\figbox}[3]{\IfFileExists{#1}{\includegraphics[width=#2]{#1}}%
  {\fbox{\parbox[c][#3][c]{\dimexpr #2-2\fboxsep-2\fboxrule}{\centering\small\itshape Figure placeholder: \detokenize{#1}}}}}

\title{Duopra: From Full-Duplex Conversation to Full-Duplex Agency}

\author{
Chiyu Chen\textsuperscript{1,2}\thanks{\;Equal contribution. Work done during Chiyu Chen’s internship at Tencent.} \quad
Pengzhou Cheng\textsuperscript{3}\footnotemark[1] \quad
Yuezhao Ma\textsuperscript{1} \quad
Xiaocheng Li\textsuperscript{1} \quad
Can Liu\textsuperscript{1} \\[1pt]
\bfseries
Haonan Hu\textsuperscript{2} \quad
Yao Yao\textsuperscript{1} \quad
Yifei Yang\textsuperscript{1} \quad
Yijie Lu\textsuperscript{1} \quad
Taolei Wang\textsuperscript{1} \quad
Yansi Li\textsuperscript{1} \\[1pt]
\bfseries
Muhuan Wu\textsuperscript{4} \quad
Chenglu Sun\textsuperscript{2}\thanks{\;Project lead.} \quad
Gongshen Liu\textsuperscript{1}\footnotemark[3] \quad
Zhuosheng Zhang\textsuperscript{1}\thanks{\;Corresponding authors.} \\[3pt]
\normalfont
\textsuperscript{1}Shanghai Jiao Tong University \quad
\textsuperscript{2}Tencent \quad
\textsuperscript{3}Shanghai University \quad
\textsuperscript{4}University of Bristol \\[2pt]
\texttt{\{chenchiyu,\,ma\_yuezhao,\,orangeade,\,liucan1013,\,yaoyao27,\,yifeiyang,}\\
\texttt{\,luyj16,\,sjtuwtl,\,yansi\_li,\,zhangzs,\,lgshen\}@sjtu.edu.cn}\\
\texttt{\{chiyuuuchen,\,rubenhnhu,\,clivesun\}@tencent.com} \quad
\texttt{chengpz@shu.edu.cn}\\[3pt]
\normalfont\small
\href{https://cychen2003.github.io/duopra/}{%
  \textcolor{black}{\faGlobe\ Project Page}%
}
\quad \textcolor{black!40}{$\vert$} \quad
\href{https://github.com/cyChen2003/duopra}{%
  \textcolor{black}{\faGithub\ GitHub}%
}
\quad \textcolor{black!40}{$\vert$} \quad
\href{https://huggingface.co/Chechen123/Duopra}{%
  \textcolor{black}{%
    \raisebox{-0.15\height}{%
      \includegraphics[height=1em]{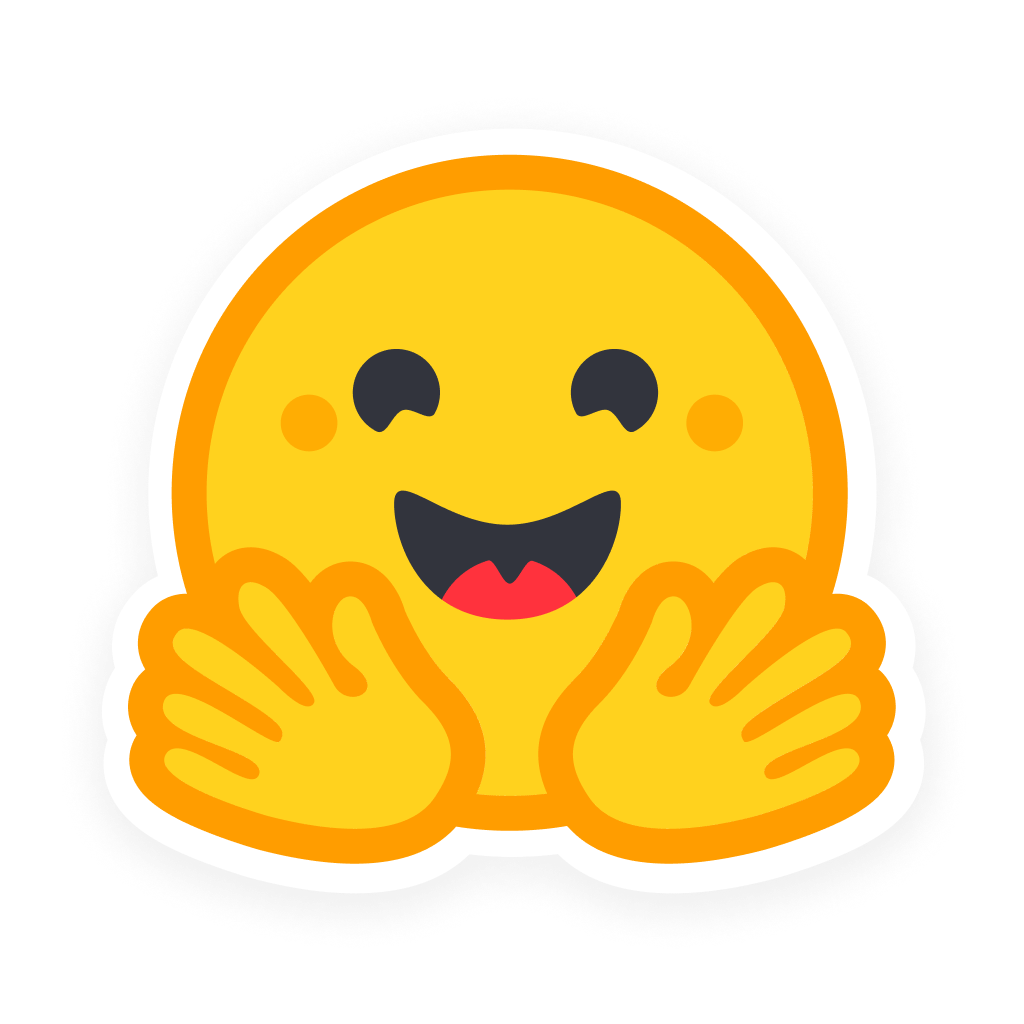}%
    }\ Hugging Face%
  }%
}
}

\begin{document}
\maketitle
\begin{abstract}
Full-duplex interaction models exhibit strong conversational agency, enabling natural, low-latency turn-taking, backchannels, and interruptions. However, completing tasks further requires task agency: timely tool invocation and argument revision following user corrections. Frontend--backend agents separate initiation from execution: the frontend decides whether and what to request, while the backend determines how. Execution can be offloaded, but unsent requests cannot be executed, and incomplete ones can misdirect the backend. On Full-Duplex-Bench v3 (FDB v3), most failures of Realtime-Venus-Audio~+ Harness in our streaming pipeline occur before a domain tool executes.
We introduce \textbf{Duopra}, a full-duplex interaction model that learns task execution alongside conversational control. Duopra treats the backend as a tool, accepting a self-contained natural-language request. Direct calls and backend delegation share an action interface and requirements for faithful initiation and correct execution. Its three-stage training combines protocol SFT, full-duplex RL rewarding timely and complete execution, and projector-only repair of talker conditioning after thinker updates. With the same backend, Pass@1 rises from 4.7\% to 44.7\%; Duopra achieves 47.7\% calling tools directly (26.0\% after SFT) and 49.3\% with a DeepSeek backend. It improves three FDB v1.0 takeover rates over the base model. These results show that task agency can coexist with key conversational agency.
\end{abstract}
\section{Introduction}
\label{sec:intro}

Full-duplex interaction models handle pauses, backchannels, turn-taking and interruptions, exhibiting \emph{conversational agency}: deciding when to listen, speak, stop or resume \citep{defossez2024moshi,cui2026minicpmo45,lin2025fullduplexbench,lin2025fdbv15}.
Tool use adds \emph{task agency}: deciding when to act, preserving the user's latest intent and completing the requested steps \citep{yao2024taubench,lin2026fdbv3}.
\emph{Full-duplex agency} requires coordinating these decisions as speech and execution results arrive.
Frontend--backend agents address task agency through a conversational frontend that sends natural-language requests to a tool-using backend \citep{venus2026realtimevenus,huang2026duplexomni,hu2026frontendbackend}.
We ask: \emph{What limits task success in frontend--backend agents, and how can a full-duplex interaction model acquire reliable task agency?}

\begin{figure*}[t]
\centering
\figbox{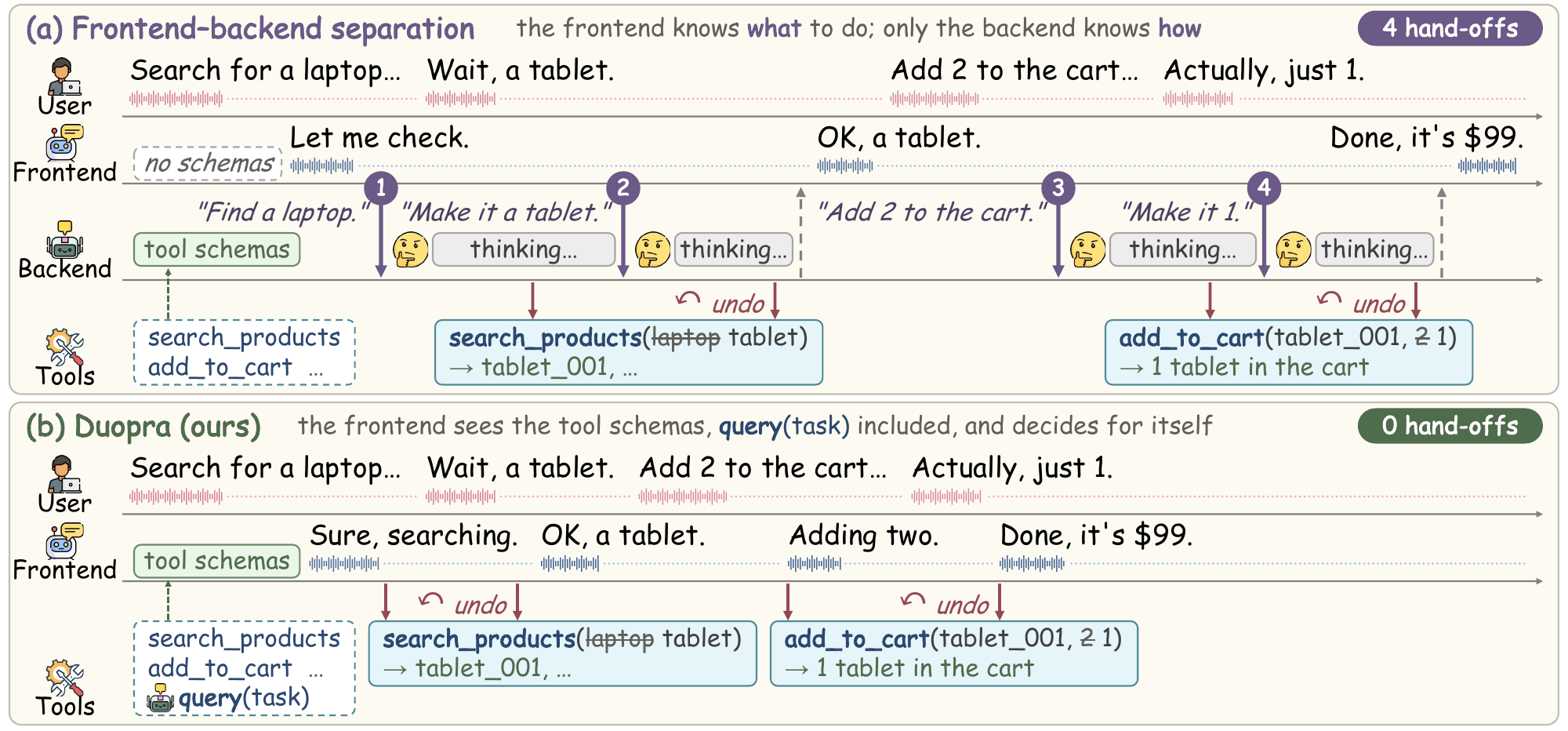}{\textwidth}{2.7in}
\caption{Frontend responsibility in an evolving FDB v3 request
(\texttt{ecommerce\_19}; schematic). Execution can be offloaded, but
initiating a faithful request remains a frontend responsibility.
The user corrects both the product (laptop to tablet) and the
quantity (two to one); \tool{add\_to\_cart} depends on the product
ID returned by \tool{search\_products}.
(a) When domain schemas are available only to the backend,
the frontend may need successive hand-offs to convey the evolving
request, making it harder to preserve corrections and tool
dependencies.
(b) With schemas visible, Duopra can issue the corrected
domain-tool calls directly without a hand-off, while
\tool{query}$(\textit{task})$ remains available.
Crossed-out values and curved arrows denote superseded requests.
Table~\ref{tab:cases}(a) reports an observed delegated success
on this request.}

\label{fig:timeline}
\end{figure*}

In this architecture, the frontend must still turn an evolving request into an executable task.
It must determine whether information is sufficient, incorporate corrections, and initiate a request that preserves the required values and steps.
Failures at this stage can prevent an otherwise capable backend from acting.
When tool schemas are available only to the backend, the frontend must also convey the request without direct access to the tools' argument requirements, potentially introducing additional clarification and execution delays.
Figure~\ref{fig:timeline} illustrates these demands with a real scenario from Full-Duplex-Bench v3 \citep[FDB v3;][]{lin2026fdbv3}: the user changes both the product and quantity, while adding the item requires a result from the preceding search.

We investigate this issue on FDB v3.
In our streaming pipeline, Realtime-Venus-Audio (RV-Audio) with its Harness, denoted RV-Audio~+ Harness, hands off requests to the backend in 11.0\% of sessions and passes 4.7\%.
Holding those hand-offs fixed caps success at this 11.0\% hand-off ceiling, even with perfect backend execution (Section~\ref{sec:diagnosis}).
The result identifies limited initiation in this configuration, rather than a general limit on delegation.

The frontend therefore needs task agency: \emph{execution can be offloaded, but initiating a faithful request remains a frontend responsibility}.
Learning this competence requires coordinating conversation with execution: the frontend must act when information is sufficient, incorporate corrections, and complete dependent steps.
Moreover, updating the language model (the \emph{thinker}) can alter the hidden states that condition the speech generator (the \emph{talker}) and degrade intelligibility.

We introduce \textbf{Duopra}, a full-duplex interaction model built on RV-Audio that controls conversation and emits private tool actions. It represents the backend as \tool{query}$(\textit{task})$, a tool with a self-contained natural-language argument, so direct calls and delegation share an action interface and execution criteria.
Its training proceeds in three stages: protocol supervised fine-tuning (SFT) teaches schema-based calling and clarification; full-duplex reinforcement learning (RL) rewards timely, complete and grounded execution; and projector-only repair realigns talker conditioning after thinker updates.

Duopra improves task completion through both direct execution and backend delegation; with the same Qwen3-8B backend, it raises Pass@1 from 4.7\% to 44.7\%.
Its takeover rates for turn-taking, pauses and interruptions on FDB v1.0 \citep{lin2025fullduplexbench} all improve over the base model, showing that task learning can coexist with turn-taking control, although spontaneous backchanneling is not retained.

Our main contributions are:
\begin{enumerate}
    \item We formalize full-duplex agency as joint conversational and task decisions, and analyze frontend initiation, argument fidelity and task completion.
    \item We introduce Duopra, a full-duplex interaction model that learns direct tool use and represents backend delegation as a natural-language-argument tool in the same action interface.
    \item We develop a three-stage procedure combining protocol SFT, execution-based full-duplex RL and projector-only repair, and evaluate task gains together with conversational and speech changes.
\end{enumerate}
\section{Related Work}
\label{sec:related}

\paragraph{Full-duplex interaction systems.}
Full-duplex models listen while speaking, requiring continuous decisions about when to respond, yield or remain silent \citep{skantze2021turntaking}.
dGSLM models textless two-channel dialogue \citep{nguyen-etal-2023-generative}, while Moshi adds an inner text monologue to parallel audio streams \citep{defossez2024moshi} and SyncLLM synchronizes generation with a real-world clock \citep{veluri-etal-2024-beyond}.
Other systems adapt pretrained LLMs to duplex speech \citep{wang2024freezeomni,hu2025salmduplex,yu2025salmonnomni,yao2025flmaudio,chen2025funaudiochat,fang2026baylingduplex,cui2026minicpmo45}.
The FDB series evaluates pauses, backchannels, turn-taking, interruptions, overlap and multi-turn dialogue \citep{lin2025fullduplexbench,lin2025fdbv15,lin2026fdbv2}, with v3 extending evaluation to tool use on real, disfluent speech \citep{lin2026fdbv3}.

\paragraph{Frontend--backend agents.}
Delegation separates ongoing conversation from asynchronous task execution.
Realtime-Venus executes each \tok{<delegate>} request in its asynchronous Harness runtime \citep{venus2026realtimevenus}, JoyAI-VL-Interaction chooses every second among speaking, silence and delegation \citep{joyai2026vlinteraction}, and DuplexOmni connects the frontend to a pluggable thinking layer \citep{huang2026duplexomni}.
\citet{hu2026frontendbackend} train a speech frontend to emit a delegation token and forward streaming ASR transcripts to a text backend.
SALMONN-duo learns when to answer directly or invoke an asynchronous backend, using cost-aware RL to balance performance and backend use \citep{yu2026salmonnduo}, while DuplexAgent-RSI improves the delegation harness from interaction traces \citep{shen2026duplexagentrsi}.
Task success in this setting depends on both the frontend's requests and the backend's execution.

\paragraph{Direct tool use and hybrid routing.}
Speech frontends can call tools directly, extending their role beyond conversation and delegation.
GPT-Realtime supports asynchronous function calling \citep{openai2025gptrealtime}; NemotronLabs VoiceChat and Qwen-Audio-3.1-Realtime also support agentic voice interaction \citep{nemotronlabs2026voicechat,bao2026qwenaudio31}.
DuplexSLA coordinates speech, language and action \citep{zhang2026duplexsla}, while SHANKS initiates tool calls during user speech \citep{chiang2025shanks}.
Qwen-Audio-Agent provides a harness supporting direct tool calls and \tool{spawn\_thinking} delegation \citep{deng2026qwenaudioagent}.
Duopra studies how the full-duplex frontend learns to initiate faithful execution, through either native calls or delegation, and analyzes the resulting task failures.

\paragraph{Post-training for spoken agents.}
Speech post-training targets semantic quality through user interactions and AI feedback \citep{wu2025aligning,lin2025alignslm}, and conversational behavior through preference optimization \citep{fang2026baylingduplex} and RL \citep{ohashi2026multifaceted,hsiao2026aspirin,li2026duplexpo}.
Qwen-Audio-3.1-Realtime uses GRPO for tool use and aligns
whether and when to speak or act \citep{bao2026qwenaudio31}.
Duopra trains task execution using time-anchored deadlines
and outcome tiers.
It then updates only the thinker--talker projector to address speech degradation, with regularization toward its original weights \citep{li2018explicit}.

Appendix~\ref{app:related} reviews broader work on tool learning, spoken evaluation, asynchronous execution and information sufficiency.
\section{Full-Duplex Agency}
\label{sec:problem}

Section~\ref{sec:intro} asks what limits task success in frontend--backend agents and how a full-duplex interaction model can acquire reliable task agency.
To make this question precise, we model the frontend as one streaming policy whose actions depend on the tools it can call.
Backend delegation and direct tool calling then become settings of one model, and information sufficiency and task success are defined independently of which component executes a call.

\paragraph{One policy for speech and action.}
User audio arrives in units $x_1,x_2,\ldots$ of length $\Delta$.
At unit~$t$, the frontend reads $x_t$ and the observations $e_t$ injected by then, which hold the schemas of the tools it can call and the results of its earlier calls, and emits $y_t=(m_t,w_t,a_t)$: a turn state $m_t\in\{\textsf{listen},\textsf{speak}\}$, text tokens $w_t$, which are voiced, and a possibly empty private action $a_t$, which is never voiced.
The policy is causal:
\begin{equation}
  \pi_\theta\left(y_t \mid x_{\le t},\, e_{\le t},\, y_{<t}\right).
  \label{eq:policy}
\end{equation}

Conversational agency decides $m_t$ and $w_t$: when to listen, speak, stop or resume through pauses, backchannels and interruptions.
Task agency decides $a_t$, which tool to call with which arguments, and the part of $w_t$ that serves the request: asking for a missing value and reporting what a call returned.
\emph{Full-duplex agency} is making both decisions well in one causal policy.

\paragraph{Tools and exposures.}
Each domain tool $f\in\Fcat$ has a JSON schema with required and optional arguments, and the backend agent $B$ is one more tool, \tool{query}$(\textit{task})$, whose argument is a self-contained natural-language task that $B$ executes with its own tools $\FB$ (Appendix~\ref{app:prompts}).
The frontend calls a tool $f$ from its set $\mathcal{T}\subseteq\Fcat\cup\{\tool{query}\}$ through one action, \tool{call}$(f,\textit{args})$; an auxiliary \tool{select} action retrieves further schemas in $\mathcal{T}$.
Request-based delegation, through which frontend--backend agents such as Realtime-Venus hand the backend a request, exposes only the backend ($\mathcal{T}=\{\tool{query}\}$, with the domain tools in $\FB$); direct calling exposes the domain tools ($\mathcal{T}=\Fcat$); and their union exposes both.
We call these three tool exposures \emph{Backend}, \emph{Direct} and \emph{Both}.
Under Both, the model chooses the route through this interface, without a separate dispatcher; execution is checked against the same required domain calls (Section~\ref{sec:diagnosis}).

\paragraph{Information sufficiency.}
A call is justified only if each of its required arguments has a source: the user's speech, in which the last correction is final (in Figure~\ref{fig:timeline}, one tablet rather than two laptops); the result of an earlier call; or the session context with the defaults it specifies.
If a required value has no source, the frontend should ask for the missing items only and act once they are given; missing optional arguments take their defaults.
The criterion is the same for both routes. A delegated task must preserve the latest known values and required steps, while identifying any values to be supplied by earlier tool results. The frontend uses the backend descriptions actually exposed in its context (Appendix~\ref{app:prompts}).

\paragraph{Success criteria.}
Each request has gold calls $\Cgold$, possibly dependent on earlier results or conditions.
\emph{Task success} requires (i)~\emph{calls}: the calls executed by the frontend or backend match $\Cgold$, with all normalized arguments correct and no call missing or extra; (ii)~\emph{grounding}: the spoken answer rests only on returned observations and covers the request; and (iii)~\emph{timeliness}: if information is sufficient, the frontend acts within $k_{\mathrm{act}}$ units after the request ends, and otherwise asks only for missing items.
\emph{Conversational control} requires (iv) that the frontend take the turn when the user finishes but not in pauses, stop within $k_{\mathrm{stop}}$ units of an interruption's onset, and continue through backchannels and background speech.

\paragraph{Initiation, execution and control.}
On FDB v3, every session holds one request that needs at least one call.
Under request-based delegation, a domain tool runs only after the frontend calls \tool{query}, so Pass@1 factors into the share of sessions with such a hand-off, which the frontend alone decides, and the pass rate of these sessions, which depends on both the \textit{task} the frontend writes and the backend that executes it.
However strong the backend, Pass@1 cannot exceed the hand-off share, which we call the \emph{hand-off ceiling}: no backend can execute a request that was never sent.
Learning to act may also affect conversation: one policy decides $(m_t,w_t)$ and $a_t$, so training it to act can change its conversational control (iv) and, because its speech is voiced from its internal states, the intelligibility of that speech.
Duopra (Section~\ref{sec:method}) implements this policy with domain tools and \tool{query} in the same action space, and Section~\ref{sec:experiments} evaluates its task success together with these conversational and speech effects.

\section{Duopra}
\label{sec:method}

\begin{figure*}[t]
\centering
\figbox{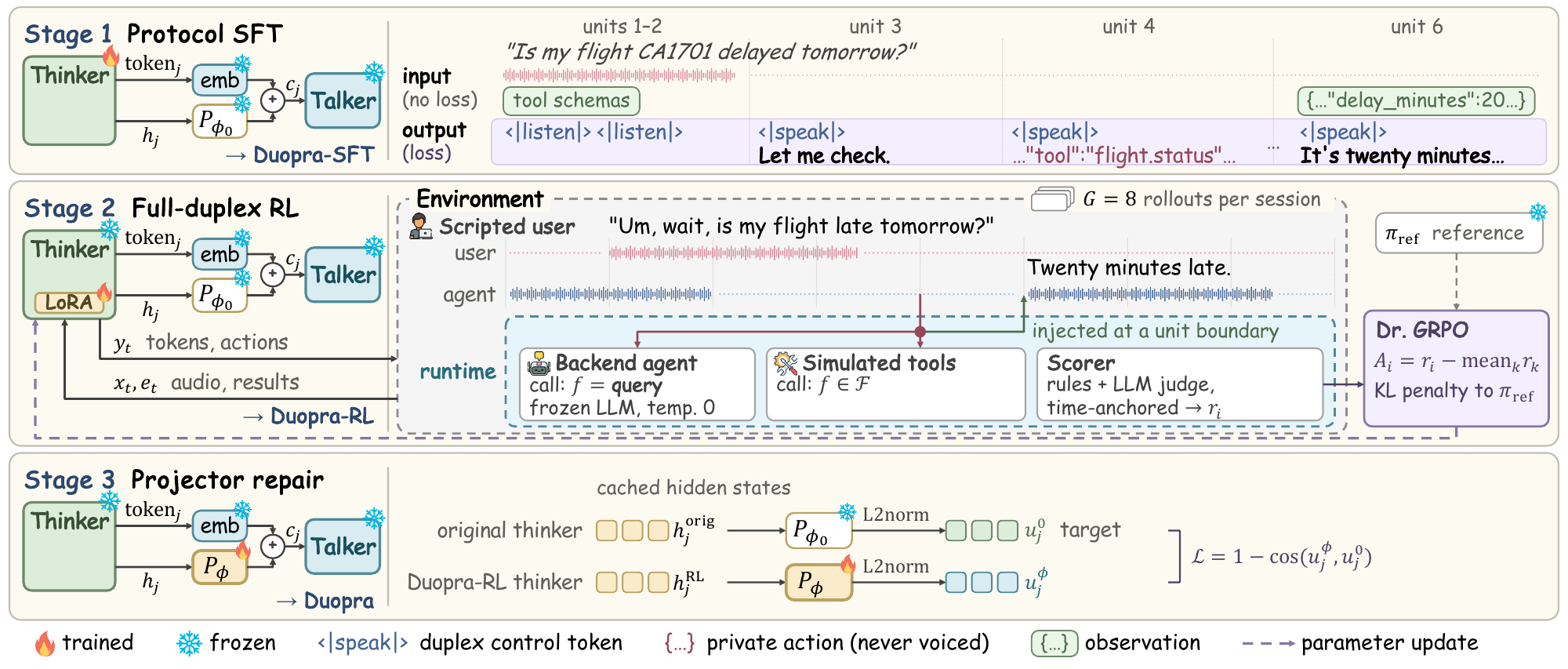}{\textwidth}{2.7in}
\caption{Duopra's three training stages on RV-Audio
($h_j$: thinker hidden state at spoken-token position $j$; $P_{\phi_0}$, $P_\phi$: base and Stage-3 projectors; $c_j$: talker conditioning).
Stage~1 trains the thinker on unit-level sessions and yields Duopra-SFT (Table~\ref{tab:sample}).
Stage~2 trains a thinker LoRA with Dr.~GRPO in the sandbox of Section~\ref{sec:rl}; Stage~2a yields Duopra-RL1 and Stage~2b yields Duopra-RL.
Stage~3 freezes the Duopra-RL thinker and trains only its projector, producing the final Duopra model.}
\label{fig:method}
\end{figure*}

Duopra is a full-duplex interaction model built on RV-Audio \citep{venus2026realtimevenus}. It replaces the base model's private free-text delegation requests with structured JSON actions. Figure~\ref{fig:method} outlines the three training stages.

\subsection{Unified Action Space}
\label{sec:model}\label{sec:action}

Duopra retains the base model's streaming loop. In each 1-s unit, the language model, or \emph{thinker}, generates duplex control tokens and public text $w_t$. The \tok{<|listen|>} and \tok{<|speak|>} tokens represent the turn state $m_t$ in Eq.~\eqref{eq:policy}. The speech generator, or \emph{talker}, voices the text conditioned on the thinker's hidden states \citep{xu2025qwen25omni}. To act within the same stream, Duopra encodes $a_t$ in Eq.~\eqref{eq:policy} as JSON in the thinker's private \tok{<delegate>} span. Duopra uses a \tool{select} action to retrieve tool schemas and a \tool{call} action to invoke an exposed tool. The runtime validates each call and inserts its result as a \tok{<backend>} observation at the first unit boundary after the result returns.

The backend is one more tool: Duopra delegates through the same \tool{call} operation by invoking \tool{query}. It supplies a natural-language task description that contains the user's latest values, requested steps, and dependencies. Appendix~\ref{app:prompts} specifies the training and evaluation interfaces.

\subsection{Stage 1: Protocol Supervised Fine-Tuning}
\label{sec:training}\label{sec:sft}

In Stage~1, we fine-tune the thinker on a mixture of tool-use demonstrations and capability-retention dialogues, represented as streams of 1-s units (Appendix~\ref{app:tasks}). The tool-use demonstrations include listening, asking for missing required values, retrieving schemas, issuing structured calls, and reporting tool results. In the flight-status example in Table~\ref{tab:sample}, the private call and the public answer occupy separate units, with the tool result arriving before the answer.

For a demonstration $d$, let $z_j^*$ be the reference thinker token at position $j$ and $H_j^*$ the stream prefix available at that position. The prefix includes user audio, observations (tool schemas and results), boundary tokens inserted by the runtime, and earlier reference outputs. The loss covers only the thinker's reference output positions $\mathcal{M}(d)$, which include public text, private actions, and duplex control tokens. We minimize
\begin{equation}
\label{eq:sft}
\mathcal L_{\mathrm{SFT}}=-\mathbb{E}_{d}
\!\left[\frac{\sum_{j\in\mathcal{M}(d)}\log\pi_\theta(z_j^*\mid H_j^*)}
{|\mathcal{M}(d)|}\right].
\end{equation}

We update the thinker's transformer layers and final normalization, keeping its embeddings and output head, the audio encoder, and speech generation modules fixed. The resulting Duopra-SFT checkpoint initializes Stage~2.

\subsection{Stage 2: Full-Duplex Reinforcement Learning}
\label{sec:rl}

\paragraph{Streaming rollouts.}
For each training task, we sample $G=8$ rollouts of the policy using the same scripted user audio and tool catalog. The runtime executes direct calls against simulated domain tools and sends Stage-2b delegations to a frozen Qwen3.8-27B backend. It inserts the resulting observations into the rollout's stream. The scorer compares the logged calls with the gold calls $\Cgold$ and reads action timing from the same log. A judge assesses spoken claims against the observations available when the claims were made, and the scorer validates the judge's cited evidence (Appendix~\ref{app:reward}).

\paragraph{Training curriculum.}
\label{sec:stage2a}\label{sec:stage2b}
Stage~2a trains on tasks involving direct calls, corrections, clarification, interruptions, and backchannels, and yields Duopra-RL1. Stage~2b adds five task families involving dependent and conditional calls, backend delegation under varying tool visibility, and recovery from failures, while replaying Stage-2a tasks; it yields Duopra-RL (Appendix~\ref{app:tasks}).

\paragraph{Reward design.}
\label{sec:reward}
Let $\operatorname{family}(\tau)$ denote the task family of rollout $\tau$.
The sets $\mathcal{D}_{\mathrm{2a}}$ and $\mathcal{D}_{\mathrm{2b}}$ contain
the Stage-2a families and the five families added in Stage~2b, respectively.
The rollout reward is
\begin{equation}
\label{eq:reward}
 r(\tau)=
 \begin{cases}
 R_{\mathrm{A}}(\tau), & \operatorname{family}(\tau)\in\mathcal{D}_{\mathrm{2a}},\\
 R_{\mathrm{B}}(\tau), & \operatorname{family}(\tau)\in\mathcal{D}_{\mathrm{2b}}.
 \end{cases}
\end{equation}

The Stage-2a reward $R_{\mathrm A}$, which also scores the replayed Stage-2a tasks in Stage~2b, combines family-specific rewards and penalties for action timing, necessary clarification, tool and argument correctness, grounded answers, and conversational control. Action timing starts at the end of the user's request, and stopping time is measured from the onset of an interruption, with deadlines ${k_{\mathrm{act}}=3}$ and ${k_{\mathrm{stop}}=2}$ units, respectively. With sufficient information, acting within $k_{\mathrm{act}}$ units earns $+0.1$. Using a value that the user has corrected incurs $-0.6$ per value.

The Stage-2b reward $R_{\mathrm B}$ scores the outcome of the full request. Complete execution with a grounded answer earns $1.2$ for action within three units of the request end, $1.1$ for action after three and within eight units, and $0.9$ for later action, before any applicable delegation cost. An honest report of partial completion earns a base reward of $0.1+0.15c$, where $c$ is the fraction of requested steps executed correctly. A fabricated result earns~$-0.95$. Domain calls from either route count toward completion, and duplicate execution incurs a penalty. Appendix~\ref{app:reward} gives the full scoring rules.

\paragraph{Policy objective.}
We use the rollout rewards to update the thinker's LoRA adapters \citep{hu2022lora} with Dr.~GRPO \citep{liu2025drgrpo}. In a retained group, $I$ indexes the scored rollouts. Rollout $i$ has reward $r_i=r(\tau_i)$ and advantage $A_i=r_i-|I|^{-1}\sum_{k\in I}r_k$. Let $\mathcal{M}(\tau_i)$ index the sampled thinker tokens in that rollout. At position $j$, $\rho_{ij}$ is the sampled token's probability under the current policy divided by its probability under the policy used for sampling. We write $\bar\rho_{ij}$ for this ratio clipped to $[1-\epsilon,1+\epsilon]$, and $K_{ij}$ for the per-token KL estimator against Duopra-SFT. We maximize
\begin{equation}
\label{eq:grpo}
\begin{split}
\mathcal J_{\mathrm{RL}}(\theta) &= \mathbb E\!\left[
 \frac{1}{|I|L}\sum_{i\in I}\sum_{j\in\mathcal{M}(\tau_i)} s_{ij}\right],\\
 s_{ij}&=\min(\rho_{ij}A_i,\bar\rho_{ij}A_i)-\beta K_{ij}.
\end{split}
\end{equation}

The expectation is over retained groups, with a fixed token normalizer $L$ and KL weight $\beta$. Appendix~\ref{app:impl} specifies group filtering, the KL estimator, and hyperparameters.

\subsection{Stage 3: Projector-Only Talker Repair}
\label{sec:projector}

Thinker training changes the hidden states used to condition the talker, which can degrade intelligibility (Section~\ref{sec:cost}).
We use the RV-Audio checkpoint before Duopra post-training as the reference and denote its projector, the base projector, by $P_{\phi_0}$. At spoken-token position $j$, the talker receives the sum of the token's text embedding and the normalized projector output.
We run the base thinker and the Duopra-RL thinker on the same stream prefix and cache their hidden states $h_j^{\mathrm{orig}}$ and
$h_j^{\mathrm{RL}}$, respectively.
The target is $u_j^0=\mathrm{L2norm}(P_{\phi_0}(h_j^{\mathrm{orig}}))$,
where $\mathrm{L2norm}$ normalizes a vector to unit $L_2$ norm.
The trainable projector gives
$u_j^\phi=\mathrm{L2norm}(P_\phi(h_j^{\mathrm{RL}}))$.
We optimize
\begin{equation}
\label{eq:projector}
\mathcal L_{\mathrm{proj}}=
\mathbb{E}_{j}\!\left[1-\cos(u_j^\phi,u_j^0)\right]
+\lambda\lVert\phi-\phi_0\rVert_2^2 .
\end{equation}

The cosine term aligns the projected Duopra-RL states with the
reference directions from RV-Audio. The L2-SP term
\citep{li2018explicit} regularizes the projector weights toward
the base weights $\phi_0$.

\section{Experiments}
\label{sec:experiments}

\begin{table*}[!t]
\centering\footnotesize
\setlength{\tabcolsep}{2.9pt}
\renewcommand{\arraystretch}{1.06}
\begin{tabular}{@{}llcccccc@{\hspace{8pt}}ccc@{}}
\toprule
\rowcolor{tablehead}& & \multicolumn{6}{c}{\begin{tabular}[c]{@{}c@{}}Task agency\\FDB v3\end{tabular}} & \multicolumn{3}{c}{\begin{tabular}[c]{@{}c@{}}Conversational agency\\FDB v1.0 TOR\end{tabular}}\\
\cmidrule(lr){3-8}\cmidrule(l){9-11}
\rowcolor{tablehead}System & Backend LLM & Call$\!\uparrow$ & Pass$|$call$\!\uparrow$ & Tool$\!\uparrow$ & Arg.$\!\uparrow$ & Resp.$\!\uparrow$ & Pass@1$\!\uparrow$ & Turn$\!\uparrow$ & Pause$\!\downarrow$ & Intr.$\!\uparrow$\\
\midrule
\multicolumn{11}{@{}l}{\textit{(a) Commercial systems, published results}}\\
GPT-Realtime                           & \na        & \na  & \na  & 87.6 & 68.0 & 79.2 & 60.0 & \na & \na & \na\\
Gemini Live 3.1                        & \na        & \na  & \na  & 81.7 & 58.8 & 71.8 & 54.0 & \na & \na & \na\\
MOSS-Live$^{m}$                        & \na        & \na  & \na  & \na  & \na  & \na  & 51.0 & \na & \na & \na\\
Gemini Live 2.5                        & \na        & \na  & \na  & 78.6 & 59.3 & 55.4 & 49.0 & \na & \na & \na\\
Cascaded                               & \na        & \na  & \na  & 80.3 & 56.2 & 60.0 & 45.0 & \na & \na & \na\\
Grok                                   & \na        & \na  & \na  & 79.7 & 54.2 & 61.7 & 43.0 & \na & \na & \na\\
Ultravox v0.7                          & \na        & \na  & \na  & 79.4 & 51.3 & 51.0 & 41.0 & \na & \na & \na\\
\midrule
\multicolumn{11}{@{}l}{\textit{(b) Open-source systems, published results}}\\
\citeauthor{hu2026frontendbackend}$^{h}$ & Qwen3-235B & \na  & \na  & 71.7 & 55.2 & 67.0 & 48.0 & \na & \na & \na\\
DuplexAgent-RSI$^{x}$                  & \na        & \na  & \na  & 97.4 & 57.0 & \na  & 47.0 & \na & \na & \na\\
RV-Audio, offline$^{o}$                & \na        & \na  & \na  & 82.0 & 52.2 & \na  & 42.0 & \na & \na & \na\\
Gander                                 & GPT-5.6    & \na  & \na  & 75.9 & 50.3 & 49.0 & 40.0 & \na & \na & \na\\
NemotronLabs VoiceChat                 & \na        & \na  & \na  & 82.5 & 42.2 & \na  & 33.0 & \na & \na & \na\\
\midrule
\multicolumn{11}{@{}l}{\textit{(c) Baselines, our streaming pipeline}}\\
MiniCPM-o~4.5 (no tools)               & none     & \na  & \na  & 0.0  & 0.0  & 28.6 & 0.0  & \textbf{86.3} & 38.7 & \textbf{99.2}\\
RV-Audio (no tools)                    & none     & \na  & \na  & 0.0  & 0.0  & 17.0 & 0.0  & 59.1 & 18.1 & 79.5\\
RV-Audio + Harness                     & Qwen3-8B & 10.3 & 45.2 & 9.5  & 5.9  & 22.9 & 4.7  & \na  & \na  & \na\\
DuplexOmni                             & Qwen3-8B & \na  & \na  & 60.0 & 27.5 & 42.0 & 17.0 & \na  & \na  & \na\\
\midrule
\multicolumn{11}{@{}l}{\textit{(d) Duopra, our streaming pipeline}}\\
\rowcolor{oursfill}Duopra-SFT (Direct) & none     & 44.7 & 58.2 & 39.2 & 30.1 & 56.8 & 26.0 & 75.1 & 9.3 & 97.0\\
\rowcolor{oursfill}Duopra-RL1 (Direct) & none     & 56.3 & 47.3 & 44.7 & 32.2 & 59.1 & 26.7 & 75.6 & 5.6 & 96.5\\
\rowcolor{oursfill}Duopra (Direct)     & none     & 81.3 & 58.6 & 76.7 & 56.2 & 66.3 & 47.7\gain{+43.0} & & & \\
\rowcolor{oursfill}Duopra (Backend)    & Qwen3-8B & \textbf{86.7} & 51.5 & \textbf{81.8} & \textbf{57.0} & \textbf{70.6} & 44.7\gain{+40.0} & & & \\
\rowcolor{oursfill}Duopra (Backend)    & DeepSeek & 77.0 & \textbf{64.1} & 71.8 & 55.0 & 69.0 & \textbf{49.3}\gain{+44.7} & \multirow{-3}{*}{75.6} & \multirow{-3}{*}{\textbf{5.4}} & \multirow{-3}{*}{91.3}\\
\bottomrule
\end{tabular}
\caption{Main results (\%). With the same backend, Duopra outperforms both frontend--backend baselines on FDB v3, and it improves all three FDB v1.0 takeover rates of RV-Audio.
Call: share of sessions with an executed domain-tool call; Pass$|$call: Pass@1 among those sessions (Pass@1 = Call $\times$ Pass$|$call); Tool: tool-name F1 (judge-free); Arg./Resp.: judged argument accuracy and response quality; Turn, Pause and Intr.: FDB v1.0 takeover rates (TOR) after the user finishes, during real pauses and after interruptions, calibrated on MiniCPM-o~4.5 (Table~\ref{tab:calibration}).
(a,b)~Published results \citep{lin2026fdbv3,mosi2026mosslive,hu2026frontendbackend,shen2026duplexagentrsi,venus2026realtimevenus,orantqing2026gander,nemotronlabs2026voicechat} use their sources' evaluation setups and judges (Table~\ref{tab:cited}); they are reference points, not a common-protocol ranking.
(c,d)~Our streaming pipeline: means over three decoding seeds (300 FDB v3 sessions per configuration; standard deviations and paired 95\% intervals in Appendix~\ref{app:results}) with the Qwen3.8-27B judge. Exposures follow Section~\ref{sec:problem}; RV-Audio + Harness and DuplexOmni use their own frontend--backend interfaces. FDB v1.0 involves no tools, so one evaluation covers the three Duopra rows.
Green subscripts: Pass@1 gain over RV-Audio + Harness. Bold: best in (c,d). \na: not reported or not applicable.
$^{m}$Tool-call success, not Pass@1. $^{h}$Tool-selection accuracy, not F1. $^{x}$On VoiceChat (49.0 on Qwen-Realtime), with exact-string argument matching. $^{o}$Offline inference with tool definitions in the prompt.}
\label{tab:main}
\end{table*}

We answer the question of Section~\ref{sec:intro} by first establishing the outcome and then explaining it:

\textbf{RQ1:} Does Duopra acquire reliable task agency?

\textbf{RQ2:} What limits task success in full-duplex interaction agents?

\textbf{RQ3:} How does training turn initiation into complete execution?

\textbf{RQ4:} Can task agency coexist with conversational agency?

\subsection{Experimental Setup}
\label{sec:setup}

\paragraph{Benchmarks.}
FDB v3 \citep{lin2026fdbv3} contains 100 disfluent English recordings and 12 simulated APIs; 66, 18 and 16 recordings require one, two and three calls, respectively.
FDB v1.0 \citep{lin2025fullduplexbench} evaluates turn-taking, pauses, backchannels and interruptions through takeover rates (TOR); FDB v1.5 and VoiceBench provide additional conversational and spoken-answer measures (Appendix~\ref{app:eval}).

\paragraph{Systems and exposures.}
We evaluate two frontend--backend baselines, RV-Audio~+ Harness and DuplexOmni, with the same Qwen3-8B backend and simulated APIs; RV-Audio and MiniCPM-o~4.5 without tools serve as controls.
Published results of commercial and open-source systems provide external reference points (Table~\ref{tab:main}).
Duopra-SFT, Duopra-RL1 and Duopra-RL denote thinker checkpoints after Stages~1, 2a and 2b; Duopra pairs the Duopra-RL thinker with the Stage-3 projector on every benchmark (Table~\ref{tab:model-configurations}).
Each Duopra checkpoint runs under one of the exposures of Section~\ref{sec:problem}: Direct (domain tools only), Backend (\tool{query} only) or Both.
The frozen backend is Qwen3-8B or DeepSeek-V4.1-Flash (DeepSeek).

\paragraph{Evaluation protocol.}
All our runs use one streaming pipeline, with three decoding seeds per FDB v3 recording (300 sessions per checkpoint and exposure).
A session passes only if its executed calls match the gold calls, with correct arguments and no missing or extra call; Pass@1 is the share of passed sessions, and response quality and factual-answer latency are measured separately.
Qwen3.8-27B judges arguments and responses.
Paired 95\% confidence intervals resample recordings with their three runs kept together and do not estimate training-run variability (Appendix~\ref{app:eval}).
Additional outcomes and cases appear in Appendix~\ref{app:results}.

\subsection{Main Results (RQ1)}
\label{sec:overall}

\paragraph{Both execution routes succeed.}
Table~\ref{tab:main} summarizes all results.
Duopra improves over both frontend--backend baselines whether it calls domain tools itself (Direct) or delegates through \tool{query} (Backend).
With the backend fixed to Qwen3-8B, replacing the frontend of RV-Audio~+ Harness with Duopra raises Pass@1 from 4.7\% to 44.7\%, and the paired 95\% interval of this gain excludes zero (Table~\ref{tab:ablation}b).
Without any inference-time backend, Duopra performs comparably to its delegated configurations, so the trained frontend both executes tasks natively and supplies executable requests to a frozen backend.

\paragraph{Published reference points.}
Published results come from other evaluation setups and judges, so they provide context rather than a ranking.
In this context, Duopra performs on par with Gemini Live~2.5 and remains behind GPT-Realtime and Gemini Live~3.1 (Table~\ref{tab:main}a,b).

\paragraph{Native execution under Both.}
When domain tools and \tool{query} are both visible, Duopra performs similarly to Direct and almost never delegates; nearly all of its successes come from its own domain calls (Appendix~\ref{app:results}).
Its task agency thus resides in the frontend and does not depend on calling the backend when the required tools are visible.

\subsection{Initiation vs.\ Execution (RQ2)}
\label{sec:diagnosis}

Having established that a frontend can acquire task agency, we locate where some baselines lose it.
Following Section~\ref{sec:problem}, Pass@1 factors into call coverage (Call), the share of sessions with an executed domain-tool call, and conditional success (Pass$|$call) among those sessions (Table~\ref{tab:main}; Figure~\ref{fig:decomp}).
A hand-off to \tool{query} need not produce a domain-tool call.

\paragraph{Initiation limits the baseline.}
RV-Audio~+ Harness hands off in only 11.0\% of sessions, so even perfect backend execution could not lift its Pass@1 above this hand-off ceiling.
With the same backend, it differs from Duopra far more in Call than in Pass$|$call (Table~\ref{tab:main}).

\paragraph{Where the gain comes from.}
Matched sessions locate the gain (Appendix~\ref{app:results}).
On the few sessions in which RV-Audio~+ Harness executes a domain call, Duopra is no more successful than the baseline.
Its gain comes from the remaining sessions, which the baseline never executes and nearly half of which Duopra passes.
Duopra thus improves Pass@1 by initiating requests that previously went unserved.

\paragraph{Effect of the exposure.}
Restricting the frontend to \tool{query} costs both accuracy and time: with Qwen3-8B, Backend yields significantly more wrong-argument sessions than Both, and with either backend LLM it delays the factual answer (Table~\ref{tab:ablation}b; Appendix~\ref{app:latency}).
These outcomes reflect both the frontend's request formulation and the backend's execution.

\paragraph{Effect of the backend.}
With Duopra fixed, replacing Qwen3-8B by DeepSeek raises conditional success but lowers call coverage, and the resulting change in Pass@1 is not significant (Table~\ref{tab:ablation}b).
A stronger backend improves the execution of initiated requests but cannot supply the requests themselves.
In frontend--backend agents, task success is therefore limited first by initiation: execution can be offloaded, but initiating a faithful request remains a frontend responsibility.

\subsection{From Initiation to Completion (RQ3)}
\label{sec:failures}

\begin{figure}[tbp]
\centering
\figbox{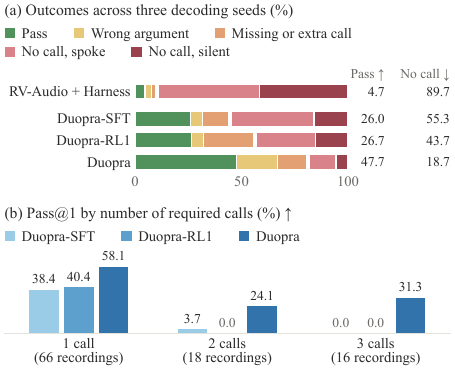}{\columnwidth}{2.50in}
\caption{Training first reduces no-call sessions and then turns calls into multi-call completions (FDB v3, 300 sessions per configuration; counts in Table~\ref{tab:fdbv3-runs}). (a)~One outcome per session for RV-Audio~+ Harness (Qwen3-8B backend) and the three Duopra checkpoints under Direct; right: shares of passed and no-call sessions. (b)~Pass@1 by the number of required calls (66, 18 and 16 recordings). $\uparrow$/$\downarrow$:~higher/lower is better.}
\label{fig:analysis}
\end{figure}

Since the bottleneck lies in initiation, we next examine how training removes it and which failures remain.

\paragraph{From calls to completions.}
Stage~2a makes the policy call tools more often, yet its Pass@1 barely changes (Figure~\ref{fig:analysis}a): calling more is not completing more.
Duopra then improves significantly over Duopra-RL1 (Table~\ref{tab:ablation}b), with the largest gains on multi-call requests, which Duopra-RL1 never completes (Figure~\ref{fig:analysis}b).
The share of sessions whose executed tool names and counts match the gold calls rises accordingly.
Training thus first increases initiation and then converts it into complete execution.

\paragraph{Residual errors.}
As no-call sessions decline, sessions with the correct tools but a wrong argument become the largest failure class (Figure~\ref{fig:analysis}a).
Selecting the required tools and preserving the user's values are separate requirements for faithful execution.
Table~\ref{tab:cases} illustrates both: a corrected request executed correctly, and an order ID in which the letter O becomes a zero.

\subsection{Conversation and Speech (RQ4)}
\label{sec:cost}

Task agency must not come at the expense of the conversational agency that makes the model full-duplex, or of the intelligibility of its speech.

\paragraph{Turn-taking control.}
Duopra improves on RV-Audio in all three FDB v1.0 takeover rates: it takes the turn more often when the user finishes or interrupts, and less often during pauses (Table~\ref{tab:main}).
The retention is behavior-specific rather than uniform, however: Duopra responds to fewer interruptions than Duopra-RL1, and none of the post-trained checkpoints produces backchannels on the tested items (Appendix~\ref{app:results}).

\paragraph{Projector repair.}
With the Duopra-RL thinker held fixed, replacing the base projector with the Stage-3 projector more than halves Chinese CER and lowers English WER on the projector-selection set (Table~\ref{tab:stage3-speech}), and it moves the talker conditioning back toward that of RV-Audio (Figure~\ref{fig:projector}).
Task agency can thus coexist with key conversational agency in one causal policy, although spontaneous backchanneling remains to be recovered.

\section{Conclusion}
\label{sec:conclusion}

This work examines how a full-duplex interaction model can move from managing conversation to carrying out tasks. Our analysis answers both parts of the question posed in Section~\ref{sec:intro}. In frontend--backend agents, task success is limited first by initiation: execution can be offloaded, but initiating a faithful request remains a frontend responsibility. We introduce Duopra, which learns to act through a unified interface for domain-tool calls and natural-language backend requests.

On FDB v3, Duopra recovers tasks left unexecuted by the evaluated baselines and improves multi-call completion beyond what more frequent invocation alone would explain. The resulting task agency coexists with improved performance on all three FDB v1.0 takeover rates, while projector-only repair mitigates speech degradation.
The loss of spontaneous backchanneling nevertheless shows that maintaining all conversational behaviors remains an open challenge.

\section*{Limitations}

\paragraph{Comparability of published results.}
We include published results for commercial systems and several open-source baselines as reference points.
Differences in evaluation setups, judges and scoring protocols limit direct comparisons with our own results.

\paragraph{Model and data coverage.}
Duopra is built on a single full-duplex interaction model and post-trained on synthetic speech from a limited set of voices.
Further evaluation is needed to determine whether the gains extend to other base models and more diverse speech conditions.

\paragraph{Future work.}
Our tool-use evaluation on FDB v3 uses prerecorded user requests and simulated APIs.
The fixed user input does not capture follow-up requests that arise from the agent's responses or tool results.
Future work could therefore evaluate Duopra in multi-turn tool-use tasks, assessing its ability to maintain task state and adapt subsequent actions as users clarify or revise their goals.

\bibliography{references}

\appendix
\raggedbottom  %
\renewcommand{\topfraction}{0.95}\renewcommand{\dbltopfraction}{0.95}
\renewcommand{\textfraction}{0.05}
\renewcommand{\floatpagefraction}{0.7}\renewcommand{\dblfloatpagefraction}{0.7}
\setcounter{topnumber}{4}\setcounter{dbltopnumber}{4}\setcounter{totalnumber}{6}
\FloatBarrier
\section{Implementation and Training Details}
\label{app:impl}
\begin{table*}[tp]
\centering
\footnotesize
\setlength{\tabcolsep}{3pt}
\renewcommand{\arraystretch}{1.10}

\begin{minipage}[t]{0.49\textwidth}
\vspace{0pt}
\begin{tabular}{@{}
>{\raggedright\arraybackslash}p{0.28\linewidth}
>{\raggedright\arraybackslash}p{0.68\linewidth}
@{}}
\toprule

\rowcolor{tablealt}
\multicolumn{2}{@{}l@{}}{\textbf{Base model: RV-Audio (9B)}} \\
\midrule
Thinker & LLM, hidden size 4096 \\
Talker & 20 layers, 768 dimensions, 12 heads \\
Audio tokens & s3tokenizer, vocabulary 6,562 \\
Vocoder & Token2wav, 24\,kHz \\
Projector $P$ & MLP, $4096\to768\to768$, 3,737,088 parameters \\

\midrule
\rowcolor{tablealt}
\multicolumn{2}{@{}l@{}}{\textbf{Stage 1: Protocol SFT}} \\
\midrule
Trainable & Thinker transformer layers and final normalization \\
Frozen & Input embedding, LM head, audio encoder, talker (incl.\ projector and $\mathrm{emb}_{\mathrm{text}}$), vocoder \\
Data & 12,800-session pool; 70\% Chinese, 60\% task-oriented, 40\% retention \\
Optimizer & AdamW; peak LR $2\times10^{-6}$, 5\% warm-up, floor $2\times10^{-7}$ \\
Training & Batch of 32 sessions; 400 updates (12,800 sessions) \\

\midrule
\rowcolor{tablealt}
\multicolumn{2}{@{}l@{}}{\textbf{Stage 2: Rollouts and Reward}} \\
\midrule
User audio & Synthesized in advance; identical across $G$ rollouts \\
Reward judge & Qwen3.8-27B; structured proof required \\
Backend & Frozen LLM, temperature 0; Qwen3.8-27B in Stage~2b \\
Objective & Dr.~GRPO, Eq.~\eqref{eq:grpo} \\
Group size $G$ & 8, sampled from the full vocabulary at temperature 1 \\
Advantage $A_i$ & Reward centered over the scored rollouts in $I$ \\

\end{tabular}
\end{minipage}\hfill
\begin{minipage}[t]{0.49\textwidth}
\vspace{0pt}
\begin{tabular}{@{}
>{\raggedright\arraybackslash}p{0.28\linewidth}
>{\raggedright\arraybackslash}p{0.68\linewidth}
@{}}
\toprule

\rowcolor{tablealt}
\multicolumn{2}{@{}l@{}}{\textbf{Stage 2: RL Optimization}} \\
\midrule
Normalizer $L$ & 2048, for surrogate and KL terms \\
Clip range & $\epsilon=0.2$ \\
KL penalty & $\beta=0.01$, k3 estimator \\
Reference & Duopra-SFT \\
Trainable & Thinker LoRA on attention and MLP; $r=32$, $\alpha=32$ \\
Learning rate & 5 warm-up steps to $3\times10^{-6}$; cosine decay to $1\times10^{-6}$ \\
Optimizer & Adam; state reset at the start of each phase \\
Groups per step & 32; at least 4 of 8 rollouts scored, with nonzero reward variance; otherwise resampled \\
Task mix (2b) & 45\% new families; 55\% Stage-2a tasks, 30\% of the latter with \tool{query} \\

\midrule
\rowcolor{tablealt}
\multicolumn{2}{@{}l@{}}{\textbf{Stage 3: Projector-Only Talker Repair}} \\
\midrule
Trainable & Projector $P_\phi$ \\
Frozen & Thinker, talker body, $\mathrm{emb}_{\mathrm{text}}$, vocoder \\
Loss & Cosine + L2-SP, Eq.~\eqref{eq:projector}, $\lambda=0.01$ \\
Training data & Cached Stage-2b thinker features; 931 sessions, 35,154 spoken-token positions \\
Optimization & 450 steps; 8,192 tokens per batch; LR $1\times10^{-4}$ \\

\end{tabular}
\end{minipage}

\par\vspace{3pt}
\noindent\rule{\textwidth}{\heavyrulewidth}

\caption{Model configuration and training settings.
Evaluation settings are given in Appendix~\ref{app:eval}.}
\label{tab:hparams}
\end{table*}

Table~\ref{tab:hparams} summarizes the model configuration and the training settings for all three stages.

\subsection{Base Model}
Duopra builds on the 9B RV-Audio model (Section~\ref{sec:method}).
Its thinker controls the duplex stream with \tok{<|listen|>}, \tok{<|speak|>}, \tok{<|chunk\_eos|>} and \tok{<|turn\_eos|>}.
The talker generates s3tokenizer audio tokens \citep[CosyVoice~2;][]{du2024cosyvoice2}, which Token2wav \citep[Step-Audio~2;][]{wu2025stepaudio2,stepfun2025stepaudio2code} converts to 24\,kHz speech.

\subsection{Stage 1: Protocol SFT}
Stage~1 initializes from RV-Audio and fine-tunes all transformer layers and the final normalization of the thinker.
The trainable and frozen components are listed in Table~\ref{tab:hparams}.
The cross-entropy loss gives equal weight to public text, private actions and duplex control tokens, averaging first over target tokens within a session and then over sessions.
Audio, observations and host-inserted tokens serve as context; the latter comprise unit delimiters and chunk endings forced by the per-unit length budget.
Training uses AdamW \citep{loshchilov2019adamw} with a peak learning rate of $2\times10^{-6}$, 5\% warm-up, a floor of $2\times10^{-7}$ and an effective batch of 32 sessions.
Duopra-SFT is the checkpoint after 400 updates, corresponding to 12,800 sessions drawn from the pool in Appendix~\ref{app:tasks}.

\subsection{Stage 2: Full-Duplex RL}

\paragraph{Rollout environment.}
Each training task defines a tool catalog, visible schemas, gold calls $\Cgold$ and a prerecorded user-audio stream.
The $G$ rollouts of a task share these inputs, including the timing of follow-up utterances such as an answer to a clarification question. The synthesis and trajectory-construction procedures are documented
in Appendix~\ref{app:tasks}.
The policy processes the stream in 1-s units while the runtime executes its private actions on simulated domain tools or on the frozen backend agent (Appendix~\ref{app:prompts}).
Each result is injected at the first unit boundary after it returns.
The runtime records executed calls and injected observations for reward computation (Appendix~\ref{app:reward}).

\paragraph{KL estimator.}
In Eq.~\eqref{eq:grpo}, we compute $K_{ij}=v_{ij}-\log v_{ij}-1$
\citep{shao2024deepseekmath}, where
$v_{ij}=\pi_{\mathrm{ref}}(z_{ij}\mid H_{ij})/
\pi_\theta(z_{ij}\mid H_{ij})$.
Here, $z_{ij}$ is the sampled token, $H_{ij}$ is its prefix, and
$\pi_{\mathrm{ref}}$ is the Duopra-SFT reference policy.

\subsection{Stage 3: Projector-Only Talker Repair}
We cache paired hidden states $(h_j^{\mathrm{orig}},h_j^{\mathrm{RL}})$ from the base thinker and the Duopra-RL thinker at each spoken-token position~$j$.
The base projector $P_{\phi_0}$ supplies the target direction
$u_j^0=\mathrm{L2norm}(P_{\phi_0}(h_j^{\mathrm{orig}}))$.
We fit the 3,737,088-parameter projector $P_\phi$ to these targets, keeping the thinker, talker body, $\mathrm{emb}_{\mathrm{text}}$ and vocoder fixed.
For a batch $\mathcal B$ of $N$ sampled spoken-token positions, the minibatch estimate of Eq.~\eqref{eq:projector} is
\begin{equation}
\begin{split}
\mathcal{L}_{\mathrm{proj}}(\phi)={}&\frac{1}{N}\sum_{j\in\mathcal B}
 \big[1-\cos(u_j^\phi,u_j^0)\big]\\
&+\lambda\,\lVert\phi-\phi_0\rVert_2^2.
\end{split}
\label{eq:projector-minibatch}
\end{equation}

The cosine term aligns projected states with their reference directions, and the L2-SP penalty \citep{li2018explicit} anchors the projector parameters to the base weights $\phi_0$.
The penalty uses the summed squared parameter differences with $\lambda=0.01$.

\subsection{Compute}
\label{app:compute}
Training runs on Kunlun P800 accelerators.
Stage~1 uses eight accelerators for 400 updates with an effective batch of 32 sessions.
Stage~2 runs full-session rollouts and LoRA updates on separate workers, with 32 groups of eight rollouts per update.
Stage~3 fits the 3.7M-parameter projector to cached features for 450 minibatch updates.

\section{Training Data and Tasks}
\label{app:tasks}
\begin{table*}[tp]
\centering
\footnotesize
\setlength{\tabcolsep}{4pt}
\renewcommand{\arraystretch}{1.1}
\begin{tabular}{@{}l>{\raggedright\arraybackslash}p{0.13\textwidth}>{\raggedright\arraybackslash}p{0.17\textwidth}>{\raggedright\arraybackslash}p{0.57\textwidth}@{}}
\toprule
\rowcolor{tablehead}Stage & Set & Size & Composition \\

1 & Training pool & 12,800 sessions &
8,951 Chinese, 3,849 English; 7,680 task-oriented, 5,120 retention (of the pool: direct calls 7.5\%, tool discovery 6.0\%, clarification 4.0\%, stopping 7.5\%, redirection 8.7\%, continuing after a backchannel 4.8\%, long reading 3.5\%, multi-turn dialogue 8.0\%, other tool tasks 10.0\%). \\
1 & Validation; test & 4,711; 4,195 sessions &
kept apart from training; they include the 16 + 16 held-out tools \\
\midrule
2a & Task pool & 7,908 tasks &
clarification-to-answer chains 34.2\%, direct calls and discovery 31.7\%, stopping and resuming 9.8\%, redirection vs.\ backchannel 7.5\%, asking first 5.9\%, answering without a tool or unavailable capability 4.5\%, corrections and cancellations 3.6\%, speech-to-tool switches 2.9\% \\
\midrule
2b & New tasks & 1,087 training; 152 validation &
training: D1 175, D2 350, D3 86, D4 368, D5 108; all required tools visible 587, some backend-only 317, all backend-only 183; 681 English, 406 Chinese. Validation: D1 30, D2 98, D4 24 \\
2b & Sampling mixture & 45\% new, 55\% Stage-2a &
new tasks and Stage-2a tasks are sampled 45:55; in 30\% of the Stage-2a samples \tool{query} is also visible, and the intended behavior there
remains a direct call \\
\midrule
3 & Cached features & 931 / 100 / 210 sessions &
The training, validation, and test sets contain
931, 100, and 210 sessions, yielding
35,154, 3,464, and 8,114 hidden-state pairs, respectively. \\
\bottomrule
\end{tabular}
\caption{Training data of each stage; the speech of Stages~1 and~2 is synthetic. Seen by Duopra-SFT:
the sessions of its 400 updates of 32 sessions. Family names: Appendix~\ref{app:tasks} and
Table~\ref{tab:families}.}
\label{tab:datacard}
\end{table*}

The training data progress from demonstrations of the action protocol to tasks requiring conversational control, planning and delegation.
Table~\ref{tab:datacard} gives the size and composition of each dataset.

\subsection{Tool Catalogs}
The initial registry contains 56 domain tools across seven domains: mobility, daily-life services, culture and recreation, personal services, factual information, environment, and facilities.
Each domain contains eight tools, such as \tool{flight.status}, \tool{parcel.status} and \tool{currency.reference}, specified by a purpose, typed required and optional arguments, and a result schema.
Stage~1 adds 16 device and sensor tools and partitions the resulting 72 definitions into 40 training, 16 validation and 16 test tools.
Stage~2a uses 52 tools: the 40 training tools, 10 additions for RL and two tools with rich results.
Stage~2b expresses 32 tool functions through 84 catalogs with distinct namespaces and argument spellings, yielding 2,688 schemas.
Of these catalogs, 64 are used for training and 20 for validation.
The coordination interfaces \tool{select} and \tool{query} are additional to these domain-tool counts.

\subsection{Stage-1 Data}
The Stage-1 pool contains 12,800 sessions, with a Chinese--English ratio of 70:30.
Table~\ref{tab:datacard} reports the task proportions.

\subsection{Stage-2a Tasks}
Stage~2a combines tool use with conversational control in 7,908 tasks, 70.5\% of which are Chinese.
The eight families cover clarification-to-answer chains; direct calls and tool discovery; clarification of missing information; responses to requests requiring no tool or an unavailable capability; corrections and cancellations; redirection and backchannels; transitions from speech to tool use; and stopping and resuming.
Their proportions are given in Table~\ref{tab:datacard}.

\subsection{Stage-2b Tasks}

\begin{table*}[tp]
\centering
\footnotesize
\setlength{\tabcolsep}{4pt}
\renewcommand{\arraystretch}{1.16}

\newcommand{\familyrule}{%
  \arrayrulecolor{black!25}%
  \specialrule{0.3pt}{4pt}{4pt}%
  \arrayrulecolor{black}%
}

\begin{tabularx}{\textwidth}{@{}
  >{\raggedright\arraybackslash}p{0.16\textwidth}
  >{\raggedright\arraybackslash}p{0.33\textwidth}
  >{\raggedright\arraybackslash}X
@{}}
\toprule
\rowcolor{tablehead}
Family & Task pattern & Expected behavior \\
\midrule

\textbf{D1}\newline Single step &
A disfluent spoken request requiring one tool call. &
Call the tool directly if visible; otherwise use
\tool{query}. Act within 3 units after the request ends. \\

\familyrule

\textbf{D2}\newline Multi-step &
Independent steps, result dependencies
(``the cheapest one''), or conditional branches. &
Complete all steps directly, or delegate a task
specifying the steps, conditions, and dependencies. \\

\familyrule

\textbf{D3}\newline Corrections &
A value is corrected or a request is abandoned
mid-sentence. &
Use the latest value and skip abandoned requests. \\

\familyrule

\textbf{D4}\newline Sufficiency &
Four variants: complete information, a missing
required value, missing optional values, or a
value obtained from a previous step. &
Ask only for missing required information;
otherwise proceed with the available values
and tool results. \\

\familyrule

\textbf{D5}\newline Recovery &
The backend requests missing information,
returns an error or partial result, or a tool
fails transiently. &
Request missing values, retry a recoverable
failure once, or report partial completion
accurately. \\

\bottomrule
\end{tabularx}

\caption{Stage-2b task families
(Section~\ref{sec:stage2b}).}
\label{tab:families}
\end{table*}

Stage~2b adds tasks for multi-step planning and recovery from backend errors and transient failures, which the Stage-2a policy rarely completes (Table~\ref{tab:families}).

\paragraph{Schemas and tool visibility.}
Each task uses a catalog with randomly namespaced tool names, such as \tool{pineridge.parcelStatus}, and paraphrased argument names.
These variants encourage the policy to interpret the delivered schema, following the schema-variation approach of SGD-X \citep{lee2022sgdx}.
All required tools are visible to the frontend in 587 training tasks (54\%), some are available exclusively through the backend in 317 (29\%), and all are available exclusively through the backend in 183 (17\%).

\section{Action Interface and Prompts}
\label{app:prompts}

\begin{table*}[tp]
\centering
\footnotesize
\setlength{\tabcolsep}{4pt}
\renewcommand{\arraystretch}{1.16}

\newcommand{\unitrule}{%
  \arrayrulecolor{black!25}%
  \specialrule{0.3pt}{3pt}{3pt}%
  \arrayrulecolor{black}%
}

\begin{tabularx}{\textwidth}{@{}
  >{\centering\arraybackslash}p{0.045\textwidth}
  >{\raggedright\arraybackslash}p{0.425\textwidth}
  >{\raggedright\arraybackslash}X
@{}}
\toprule
\rowcolor{tablehead}
Unit & Input: user audio and observations & Thinker output \\
\midrule

\textbf{1} &
\textit{Is my flight CA1701 \dots}
\par\smallskip
\textit{Tool schema:}
\par
\tok{<backend>}
\texttt{\{"tools":[\{}%
\texttt{"name":"flight.status",}\newline
\texttt{"parameters":\{"required":}\newline
\texttt{["flight","date"],...\},}\newline
\texttt{"version":"2",...\},...\}]\}}
\tok{</backend>}
&
\tok{<|listen|>} \\

\unitrule

\textbf{2} &
\textit{\dots delayed tomorrow?}
&
\tok{<|listen|>} \\

\unitrule

\textbf{3} &
\textit{(silence)}
&
\tok{<|speak|>}
\textbf{Let me check.}
\tok{<|chunk\_eos|>} \\

\unitrule

\textbf{4} &
\textit{(silence)}
&
\tok{<|speak|>}
\tok{<delegate>}
\newline
\texttt{\{"op":"call","call\_id":"c1",}
\newline
\texttt{"tool":"flight.status",}
\newline
\texttt{"arguments":\{"flight":"CA1701",}
\newline
\texttt{"date":"2026-11-06"\},}
\newline
\texttt{"schema\_version":"2"\}}
\newline
\tok{</delegate>}
\tok{<|chunk\_eos|>} \\

\unitrule

\textbf{5} &
\textit{(silence)}
&
\tok{<|speak|>}
\tok{<|turn\_eos|>}
\tok{<|chunk\_eos|>} \\

\unitrule

\textbf{6} &
\textit{(silence)}
\par\smallskip
\textit{Tool result:}
\par
\tok{<backend>}
\newline
\texttt{\{"call\_id":"c1",}
\newline
\texttt{"payload":\{"result":\{}
\newline
\texttt{"delay\_minutes":20,}
\newline
\texttt{"status":"delayed"\},...\}\}}
\newline
\tok{</backend>}
&
\tok{<|speak|>}
\textbf{It's twenty minutes}
\tok{<|chunk\_eos|>} \\

\unitrule

\textbf{7} &
\textit{(silence)}
&
\tok{<|speak|>}
\textbf{late.}
\tok{<|turn\_eos|>}
\tok{<|chunk\_eos|>} \\

\bottomrule
\end{tabularx}

\caption{
Illustrative Stage-1 training session from
Figure~\ref{fig:method}, shown in 1-s units.
}
\label{tab:sample}

\vspace{2pt}
\begin{minipage}{0.98\textwidth}
\raggedright\footnotesize
\textit{Notes.}
User audio is shown as text.
Bold text is spoken; private
\tok{<delegate>} actions are not.
Tool schemas and results are abbreviated for display;
the original observations use canonical JSON.
The session date is 2026-11-05, making
``tomorrow'' 2026-11-06.
Training loss applies to thinker-generated tokens,
excluding input, observations and forced boundaries.
\end{minipage}
\end{table*}

\begin{table*}[tp]
\centering\footnotesize
\setlength{\tabcolsep}{3pt}
\renewcommand{\arraystretch}{1.14}
\newcommand{\duoprapromptsep}{%
  \arrayrulecolor{black!22}%
  \specialrule{0.32pt}{3.5pt}{3.5pt}%
  \arrayrulecolor{black}%
}
\begin{tabular}{@{}
  >{\raggedright\arraybackslash\bfseries}p{0.155\textwidth}
  @{\hspace{9pt}}
  >{\raggedright\arraybackslash}p{0.805\textwidth}
@{}}
\toprule
\rowcolor{tablehead} Field & Verbatim content \\
\midrule
\textbf{Thinker system prefix} &
\texttt{Streaming Omni Conversation.} \\
\midrule
\rowcolor{tablealt}\multicolumn{2}{@{}l@{}}{\textbf{Backend tool \tool{assistant.query} in training}} \\
\duoprapromptsep
Description &
Delegate a complete, self-contained task to a frozen background assistant. This tool has exactly ONE argument: task, a string. Put the actual complete request and all known facts inside that string. Do not pass separate city, date, account, period or other business-tool fields to assistant.query. Use a visible direct tool for a simple supported request.\par\smallskip For several goals, result dependencies, conditions, or a capability available only in the background, send all active steps, constraints and already-heard values in task. Required values may come from the user, explicit session context, defaults, or a named preceding result. Ask only for an essential required value that has none of these sources; optional fields may use their defaults. Keep final corrections and exclude abandoned requests.\par\smallskip The background cannot hear this conversation. No transcript, previous select or hidden facts are added to task. Only registered capabilities and session permissions apply.\par\smallskip After done, need\_info or error, answer honestly from delivered observations, ask for genuinely missing information, or retry at most once after an error. Do not speak routing fields or private JSON. \\
\duoprapromptsep
Argument &
\texttt{task}, a string of 1--1,600 characters: Complete active request, all known values, ordered steps, conditions and result dependencies. \\
\duoprapromptsep
Result &
\texttt{status} (\textsf{done}, \textsf{need\_info} or \textsf{error}), \texttt{answer}, \texttt{steps}, \texttt{missing\_information}; the backend LLM writes all but \texttt{steps}, which the runtime fills from its call log. \\
\midrule
\rowcolor{tablealt}\multicolumn{2}{@{}l@{}}{\textbf{Backend tool \tool{assistant.query} in the FDB v3 adapter (Qwen3-8B)}} \\
\duoprapromptsep
Description &
Ask the Qwen3-8B task assistant to solve a request using the available tools. Useful for multi-step tasks, combining different tool capabilities, or reasoning about returned results. Pass the complete already-heard query and relevant known facts. The assistant can return a grounded answer or ask for missing information. \\
\duoprapromptsep
Argument &
\texttt{query}, a string of 1--1,000 characters: Already-heard user request and relevant known facts. \\
\midrule
\rowcolor{tablealt}\multicolumn{2}{@{}l@{}}{\textbf{Backend agent at evaluation}} \\
\duoprapromptsep
Instruction &
Complete the delegated task using only the provided functions and supplied conversation facts. Do not invent required arguments. Ask a concise clarification if information is missing. Use returned values as evidence; never claim a call succeeded if it failed. Call additional functions when the task needs multiple steps. Give a concise final answer in the language of the request. All functions are deterministic benchmark simulations. \\
\bottomrule
\end{tabular}
\caption{Verbatim system prefix, backend-tool schema text and evaluation-backend instruction. Tool schemas enter the thinker's context as observations. The backend tool is registered as \tool{assistant.query} and abbreviated as \tool{query} throughout the paper. Appendix~\ref{app:prompts} describes the tool context and the backend's inputs during training and evaluation.}
\label{tab:prompts}
\end{table*}

Table~\ref{tab:sample} illustrates the interface in a flight-status request.
The thinker listens for two units, says ``Let me check,'' calls \tool{flight.status}, ends its turn, and reports the result after the observation arrives at unit~6.
Table~\ref{tab:prompts} reproduces the system prefix and tool descriptions.

\subsection{Actions and Observations}
Duopra uses \tok{<delegate>}\,\dots\,\tok{</delegate>} for private JSON actions and \tok{<backend>}\,\dots\,\tok{</backend>} for returned observations, extending the stream format of RV-Audio (Section~\ref{sec:action}).
Both direct tool calls and backend requests use this same interface.
A \tool{select} action contains \texttt{\{op, call\_id, query\}} and retrieves the schemas of the three most relevant tools; its \texttt{query} field is the schema-search string.
A \tool{call} action contains \texttt{\{op, call\_id, tool, arguments, schema\_version\}}, with a JSON object for \texttt{arguments} and an optional \texttt{schema\_version}.
The runtime validates each call against a schema delivered at initialization or retrieved by \tool{select}, returning a structured error for invalid actions.
Each observation carries the action's \texttt{call\_id} and enters the stream at the first unit boundary after completion.
The frontend then uses the delivered result to formulate its response.
Private actions remain in the thinker's autoregressive history, while a per-unit mask restricts the talker's conditioning to public speech.

\subsection{Tool Context and the Backend Tool}
Training and evaluation retain the native system prefix, ``Streaming Omni Conversation.''
Tool schemas enter as observations and supply the tool-specific instructions, including those governing delegation.
The initial schemas form the first observation as canonical JSON (Table~\ref{tab:sample}, unit~1), and \tool{select} retrieves further schemas.
Under Direct, the frontend receives the visible domain-tool schemas; Both adds \tool{query}; Backend exposes \tool{query} and assigns domain tools to the backend (Appendix~\ref{app:eval}).
Training tasks select schemas from their catalogs; FDB v3 uses its 12 benchmark APIs.

The backend tool is registered as \tool{assistant.query}, abbreviated to \tool{query} in the paper.
It accepts one natural-language string, named \texttt{task} in training and \texttt{query} in the FDB v3 adapter.
Its training description specifies the information-sufficiency criterion and the situations suited to delegation, while a capability list states the information required by each backend function.
The concrete backend tool names and schemas remain in the backend's catalog $\FB$.
For example, one request is ``Retrieve the autumn-term dates for Riverside Upper School District in Nanjing for academic year 2026, then check whether the returned start date is a public holiday in China.''
This request uses \tool{school.term\_dates} followed by \tool{holiday.calendar}, explicitly supplying the first result as the second call's input.
Such task descriptions convey known values, ordered steps, conditions and dependencies, including values to be obtained by earlier steps.

The backend is a frozen LLM decoded at temperature~0: Qwen3.8-27B for Stage-2b rollouts and Qwen3-8B or DeepSeek-V4.1-Flash for evaluation.
It executes the request with its own tools $\FB$, which can extend beyond the frontend's domain tools $\Fcat$.
During training, its input consists of the \texttt{task} string and its tool schemas.
Its response is a JSON object with three fields: \texttt{status} ($\textsf{done}$, $\textsf{need\_info}$ or $\textsf{error}$), \texttt{answer} and \texttt{missing\_information}.
The last field lists at least one missing fact for $\textsf{need\_info}$ and is empty for $\textsf{done}$; malformed responses count as failed tasks.
The runtime augments this response with \texttt{steps} from the executed-call log.
At evaluation, the backend receives the instruction in Table~\ref{tab:prompts}, the current request and earlier request--reply pairs from the session, and may execute at most six tool calls per request.
The adapter returns its response as the corresponding observation.
Thus, the frontend's task descriptions provide the backend's access to the user's request in both settings.

\section{Reward Specification}
\label{app:reward}
\begin{table*}[tp]
\centering
\footnotesize
\setlength{\tabcolsep}{4pt}
\renewcommand{\arraystretch}{1.14}
\newcommand{\duoprarewardsep}{%
  \arrayrulecolor{black!22}%
  \specialrule{0.32pt}{3pt}{3pt}%
  \arrayrulecolor{black}%
}
\begin{tabular}{@{}
  >{\raggedright\arraybackslash\bfseries}p{0.19\textwidth}
  @{\hspace{10pt}}
  >{\raggedright\arraybackslash}p{0.765\textwidth}
@{}}
\toprule
\rowcolor{tablehead} Component & Reward rule \\
\midrule
Action decision &
Information is sufficient and the frontend acts within 3 units after the user finishes: $+0.1$.
Information is sufficient but the frontend does not act: staying silent $-1$; speaking without acting $-0.8$; asking an unnecessary question $-0.6$; each of the three additionally incurs $-0.3$ for the missing call. \\
\duoprarewardsep
Sufficiency &
A required argument is missing and the frontend asks only for the missing items: $+0.3$.
Asking for optional or already given information: $-0.15$ per item.
Acting although a required argument is missing: $-0.3$.
Inventing a value for the user: $-0.6$. \\
\duoprarewardsep
Step coverage &
$+0.2\,c$, where $c$ is the fraction of requested steps that are executed correctly.
Including a task the user abandoned: $-0.3$. \\
\duoprarewardsep
Argument fidelity &
$+0.3\,c$; a step counts toward $c$ only if all its values are correct.
Each value from before a correction: $-0.6$. \\
\duoprarewardsep
End-to-end &
The executed calls (direct and backend) match $\Cgold$ exactly in tools and arguments: $+0.2$. \\
\duoprarewardsep
Result readout &
Complete readout $+0.4$; partial readout $+0.2$.
Reporting a result before its observation arrives is marked as fabrication;
the applicable cap and outcome mapping are specified separately for
each task family in Appendix~\ref{app:reward}. \\
\duoprarewardsep
Backend status &
After \textsf{need\_info}, a brief follow-up question to the user and a new \tool{query}: $+0.1$.
After \textsf{error}, a successful retry: $+0.2$.
Claiming full completion when only part was completed: $-0.8$. \\
\duoprarewardsep
Response timing &
Speaking within 3 units after the user finishes: no penalty; after 4--6 units: $-0.2$; after more than 6 units or never: $-1$.
Taking the turn during a mid-utterance pause of ${\ge}0.7$\,s: $-0.3$. \\
\duoprarewardsep
Interruption control &
Stopping within 2 units after the user starts speaking: no penalty; within 3--4 units: $-0.3$; within 5--6 units: $-0.6$; later: $-1$.
After a backchannel or noise: pausing for more than 1 unit $-0.5$; ending the content $-1$. \\
\duoprarewardsep
Routing and duplication &
Delegating the same request twice, or both calling directly and delegating it: $-0.2$.
Delegating a single-step task whose tool is visible, when the delegation succeeds: $-0.05$ (latency cost). \\
\bottomrule
\end{tabular}
\caption{Reward components and checks.
Times are measured in 1-s audio units from each rule's stated anchor. Step coverage $c$ counts correctly executed steps from either the frontend or the backend. Stage-2a task families use $R_{\mathrm A}$; the five new Stage-2b families use the outcome mapping $R_{\mathrm B}$ in Appendix~\ref{app:reward}.}
\label{tab:reward}
\end{table*}

\begin{table*}[tp]
\centering\scriptsize
\setlength{\tabcolsep}{3pt}
\renewcommand{\arraystretch}{1.10}
\begin{tabular}{@{}
  >{\raggedright\arraybackslash}p{0.476\textwidth}
  @{\hspace{8pt}{\color{black!22}\vrule width 0.35pt}\hspace{8pt}}
  >{\raggedright\arraybackslash}p{0.476\textwidth}
@{}}
\toprule
\rowcolor{tablehead}
\multicolumn{2}{@{}l@{}}{\textbf{Reward-judge instruction (Qwen3.8-27B; verbatim)}} \\
\midrule
\rowcolor{tablealt}
\textbf{Part I: Assess claims and completion} &
\textbf{Part II: Ground evidence and return labels} \\
\midrule
Judge ONLY these public words using ONLY supplied consumed\_evidence. No future result exists in this request. covered\_goals is an array of zero-based requested\_goals indices whose actual result is correctly stated, or whose observed service failure is honestly disclosed. Promises, questions and repeating requests cover NO goals. Check every factual claim separately, especially negation, quantities, dates and identifiers.\par\smallskip Judge factual correctness separately from speaking fluency: a grammatical classifier error or filler does not change an otherwise explicit numerical value and unit. Do not label that alone as fabricated. An incomplete answer is not automatically a contradiction: stating only August when the date is in August is incomplete (does not cover a full-date goal), but not fabricated. A different month is contradictory. Similarly, six twelve without AM/PM is a 12-hour clock component, not an explicit 06:12 claim; do not mark it fabricated solely because the source uses 18:12. An explicit contradictory AM/PM is wrong. Digit-by-digit place numbers (Shenzhen three nine five versus Shenzhen 395) denote the same entity; different digits or years remain wrong.\par\smallskip fabricated=true for ANY asserted factual result absent from or contradictory to consumed\_evidence. An answer spoken without any evidence is fabricated even if coincidentally true. A refusal or an unsupported claim that a tool is unavailable is false\_refusal, NOT a fabricated business result. This includes claiming that a lookup failed, a record was unavailable, or nothing was found when no consumed error/empty-result observation supports that claim. Lack of evidence is NOT evidence that a lookup failed. With consumed\_evidence=[], saying I have not checked yet is honest uncertainty; saying I checked but the record was unavailable is false\_refusal=true. An actually consumed matching error justifies the refusal. Saying I cannot confirm, I will search for a tool, or the tool is unavailable does not assert a business answer. Set fabricated only for asserted business facts (including vague flight-on-time, open/closed, or all-clear claims). An observed error may justify an honest failure disclosure but NEVER proves that the business goal succeeded.\par\smallskip unfinished\_disclosures is an array of [goal\_index,exact\_quote] explicitly saying that a requested step was not completed, could not be checked, or remains unresolved. Promises and merely not mentioning a step are NOT disclosures. A general statement that the whole background task failed can disclose all its goals, but earns no business completion credit. Do not invent a statement about an omitted goal. assistant.query status done is only a backend claim; actual business results, not completion prose, establish truth. false\_all\_done=true when the speaker claims all steps succeeded but evidence is incomplete or failed. allowed\_goal\_sources explicitly lists the only real source indices for each goal.\par\smallskip An empty list means NO evidence for that goal; never infer its result from another goal, the request, or the backend completion claim. &
For EVERY covered\_goals entry supply exactly one goal\_proofs entry [0,0,"result","exact public words answering that goal"], ordered as goal\_index,record\_index,kind,quote. kind must be result or observed\_failure. The record\_index must occur in allowed\_goal\_sources for that goal and kind; quoted words must really answer that specific goal. A supported result for one goal cannot support another.\par\smallskip Unsupported assertions still set fabricated=true, but do not belong in covered\_goals or goal\_proofs. For each stated RESULT number, complete calendar date, clock time or identifier, numeric\_claims MUST contain ["exact numerical words from public\_text",0,["payload","result","field"]], ordered as quote,record\_index,path. quote must be the exact spoken quantity INCLUDING its unit and any classifier, WHOLE stated date including its year WHEN SPOKEN, WHOLE clock time including any AM/PM words, or WHOLE identifier, not a corrected value. Include trailing currency qualifications in the same quote: dollars in CNY explicitly names CNY; do not shorten it to dollars. Compare the explicit currency independently of the number. A month/day without a spoken year is only a month/day claim; cite it verbatim and NEVER add an inferred year. If a year is split across chunks such as 20 31, retain those exact spaces in the quote. Use separate claims for separate fields.\par\smallskip path starts in consumed\_evidence[record\_index]; numeric\_sources is the complete catalog of numeric/date/time/ID fields actually available NOW. Copy record\_index and path EXACTLY from that catalog; paths to result fields begin with payload,result. Currency names, city names, ordinary place names, and demonstratives such as that date or this number are NOT numeric claims. A demonstrative contains no stated value; numeric\_claims must omit it. If the catalog is empty, do not invent source indices.\par\smallskip A promise repeating an input ID/date has numeric\_claims=[] even when the real call will occur later. Unsupported RESULT assertions instead use null,[] and set fabricated=true. Repeating an input in a promise does not assert a result. A repeated input identifier (such as a subscription ID) is not a newly returned amount. If it is cited, its proof MUST point to ["arguments","actual\_argument\_field"] of the matching real call, NEVER to an amount or date field. Repeating input values in a promise or question is not a result claim; numeric\_claims may be empty for such speech.\par\smallskip include contradictory claims too, pointing to the field they contradict. If no corresponding numeric result field exists, set record\_index to null and path to []; never invent a field or record. Do NOT round counts, dates, IDs or quantities. The Python caller independently parses quote and compares to that field. private\_metadata\_spoken flags internal tool names, JSON and routing/schema instructions, not ordinary nouns. reason explains the comparison in at most 12 words; no other response fields. Use compact JSON on one line, without indentation or line breaks. Never repeat the whole input. \\
\bottomrule
\end{tabular}
\caption{Verbatim reward-judge instruction (read the left column before the right; part labels and paragraph breaks are for display). The input includes public text, preceding observations (consumed\_evidence), the observations supporting each requested goal (allowed\_goal\_sources) and their numeric fields (numeric\_sources). The returned labels, supporting quotes and numeric claims are validated before reward computation (Appendix~\ref{app:reward}).}
\label{tab:judge}
\end{table*}

Table~\ref{tab:reward} lists the reward components used in Eq.~\eqref{eq:reward}. Table~\ref{tab:judge} gives the instructions for the reward judge.

\subsection{Scoring Evidence}
The scorer combines runtime checks with semantic labels from Qwen3.8-27B \citep{qwen2026qwen38}.
Runtime logs and the unit timeline establish action validity, executed tools and arguments, repeated calls, and the timing of speech and actions.
Argument grounding is checked against the user's latest stated values, earlier tool results and session context.
The LLM judge evaluates claims in the public response using the observations available when those claims were made.
Its inputs include the public text, observations permitted for each requested goal, and a catalog of numeric result fields; its instruction is reproduced in Table~\ref{tab:judge}.

The judge returns structured labels for covered goals, fabricated results, false refusals, false claims of completion and spoken private metadata.
Each covered goal requires a proof containing the goal index, observation index, evidence type (result or observed failure) and a verbatim quote.
The output also identifies disclosures of unfinished steps and links stated result numbers, dates, times and identifiers to their supporting evidence fields.
The accompanying reason is limited to twelve words.
The scorer validates the output schema, verifies that quotes occur in the public text, and checks that each cited observation supports the specified goal and precedes the statement.
It then parses the quoted numeric values and compares them with the cited fields after normalizing Chinese and Arabic numerals, spoken digit sequences, time expressions and full-width characters.
For example, ``three nine five'' can mean 395, and ``six twelve'' can express the clock component of 18:12.

Validated labels determine result-readout credit, fabrication and completion status; the deterministic outcome mapping then assigns the scalar reward.
Invalid judgments leave a rollout unscored and invoke the group-filtering procedure in Appendix~\ref{app:impl}.
The remaining reward components follow runtime rules, with semantic judgments used for value normalization and for distinguishing continued content from a new response after interruption.

\subsection{Outcome Tiers}
For Stage-2a task families, $R_{\mathrm A}$ sums the applicable
components in Table~\ref{tab:reward}, with clipping to $[-1,1.2]$
and any applicable family-specific caps.

For the five new Stage-2b task families, $R_{\mathrm B}$ uses the outcome scores in Section~\ref{sec:reward}. When a call is required and none is executed, the score is $-0.85$ if the policy speaks and $-0.90$ if it stays silent.

\subsection{Timing and Backend Responsibility}
Timing rewards use 1-s units on the input-audio timeline.
Interruption, cancellation and correction deadlines start at the onset of the corresponding user utterance; response and action deadlines start at the end of the request.
The acting and stopping deadlines are $k_{\mathrm{act}}=3$ and $k_{\mathrm{stop}}=2$ units, respectively (Section~\ref{sec:problem}).
The control reward penalizes both delayed stops after interruptions and unnecessary stops after backchannels or background noise.

Unexpected backend failures are filtered when the delegated task is faithful and sufficient and the frontend's own calls and claims are correct.
This filtering separates backend execution faults from the frontend decisions optimized by RL.
Scripted errors, requests for missing information and partial completions in family D5 remain part of training, where the frontend must recover or report the outcome accurately.

\subsection{Routing and Duplicate Work}
For multi-step tasks, execution credit is determined by the fraction $c$ of steps completed correctly, aggregating direct calls and calls executed by the backend.
The route is selected by the policy through its action interface.
Two additional terms encourage efficient execution: a $-0.2$ penalty for duplicated work and a $-0.05$ latency cost for successful delegation of a single-step request whose tool is visible.
For Stage-2b task families, these terms enter the outcome mapping: the latency cost is deducted from a complete-success score, whereas duplicated work or a protocol fault places the rollout in a lower outcome tier with bounded deductions.

\section{Evaluation Protocol}
\label{app:eval}

\subsection{Tool Exposures, Backends and Models}
We evaluate three tool exposures: Direct provides the frontend with the domain tools, Both adds \tool{query}, and Backend provides \tool{query} while assigning the domain tools to the backend's catalog $\FB$.
The evaluation backends are Qwen3-8B and DeepSeek-V4.1-Flash; the latter has 552B backbone parameters and 16B active parameters per decoded token \citep{deepseekai2026deepseekv41flash}.
Both backends are frozen, use temperature~0 and execute at most six tool calls per \tool{query} request.
Stage-2b training uses Qwen3.8-27B as the backend.
The FDB v3 adapter names the backend tool's single natural-language argument \texttt{query}, corresponding to \texttt{task} in the training schema.

Table~\ref{tab:model-configurations} lists the thinker and projector used in each evaluation.

\begin{table}[tbp]
\centering\footnotesize
\setlength{\tabcolsep}{3pt}
\renewcommand{\arraystretch}{1.1}
\begin{tabular}{ll>{\raggedright\arraybackslash}p{0.45\columnwidth}}
\toprule
\rowcolor{tablehead}Model & Thinker & Projector \\
\midrule
Duopra-SFT & after Stage~1 & base \\
\dmid & after Stage~2a & base; used on FDB v1.0, FDB v1.5, VoiceBench and FDB v3 (Direct) \\
Duopra-RL & after Stage~2b & base; used for the projector drift analysis (Figure~\ref{fig:projector}b) \\
Duopra & Duopra-RL  &
Stage-3 projector (all benchmarks)\\
\bottomrule
\end{tabular}
\caption{Thinker and projector configurations used in evaluation.
All results labelled Duopra on FDB v3, FDB v1.0, FDB v1.5 and
VoiceBench use the Stage-3 projector.
Projector ablations in Appendix~\ref{app:projector} are specified separately.}
\label{tab:model-configurations}
\end{table}

\subsection{FDB v3 Metrics}
FDB v3 \citep{lin2026fdbv3} contains 100 English recordings from 12 speakers, including non-native accents, fillers, pauses and self-corrections.
The requests use 12 simulated APIs and are followed by 30\,s of ambient noise.
There are 36 easy, 34 medium and 30 hard requests; 66 require one call, 18 require two and 16 require three.
We run each configuration with three decoding seeds, yielding 300 sessions, and apply the benchmark's scoring with Qwen3.8-27B at temperature~0 as the argument and response judge.
The original benchmark uses GPT-4o \citep{openai2024gpt4o}.

\paragraph{Pass@1.}
A session passes when its execution log contains every gold call with correct arguments and no additional call, counting calls made by either the frontend or the backend.
This criterion includes calls made before a user correction: executing a superseded call makes the session fail.
The Call measure in Table~\ref{tab:main} and Figure~\ref{fig:decomp} counts sessions with an executed domain-tool call; \tool{query} records a hand-off to the backend.

\paragraph{Tool, argument and response scores.}
Tool is the F1 between the multisets of executed and gold tool names, and Arg. is the judge's argument accuracy.
Both scores are averaged over all sessions, with zero assigned to sessions without an executed call.
Resp. is the judged quality of the spoken response, averaged over sessions with a scored response, following the benchmark's aggregation.
Take is the fraction of sessions with a recorded response.

\paragraph{Outcome classes.}
We assign each session to one class in the following order: pass; no executed call; missing call; extra call; correct tools with a wrong argument.
When available, the no-call class is further divided into sessions with and without speech (Table~\ref{tab:fdbv3-runs}).
All outcome rates use the full set of sessions as their denominator.
The pass and wrong-argument classes together give the share whose executed tool names and counts match the gold calls, the judge-independent measure used in Section~\ref{sec:failures}.

\subsection{Statistics and Latency}
\paragraph{Paired comparisons.}
We estimate 95\% confidence intervals for differences using 20,000 paired bootstrap resamples of the 100 recordings.
Each resample draws recordings with replacement and retains all three decoding seeds of each selected recording in both configurations; the interval uses the 2.5th and 97.5th percentiles of the resulting differences.
For an individual configuration, we report the mean and sample standard deviation over its three seeds.
Differences are calculated from unrounded session counts.

\paragraph{Factual-answer latency.}
We compare latency between Duopra's tool exposures for each backend.
The benchmark scorer measures the time until the agent delivers the factual answer \citep{lin2026fdbv3}; this measure includes the serving runtime and tool execution, rather than measuring action-initiation timing.
We compute this measure for each seed and report its mean and sample standard deviation; the comparison appears in Appendix~\ref{app:latency}.

\subsection{Baselines}
\paragraph{RV-Audio~+ Harness.}
We run the streaming RV-Audio model with its Harness, using Qwen3-8B as the backend and the same simulated tools as Duopra.
The published FDB v3 score of 42.0\% \citep[Table~7]{venus2026realtimevenus} uses offline inference with tool definitions in the prompt, according to a repository collaborator;\footnote{Reply in a public issue of the repository, November 2026.} our evaluation uses the streaming delegation interface.

\paragraph{DuplexOmni.}
DuplexOmni \citep{huang2026duplexomni} uses Qwen3-8B as its pluggable thinking layer with the same simulated tools.
Its three runs pass 15, 17 and 19 of the 100 requests (17.0\,$\pm$\,2.0\%).
Across these runs, it passes 31 of 108 easy, 15 of 102 medium and 5 of 90 hard sessions (28.7\,$\pm$\,3.2\%, 14.7\,$\pm$\,2.9\% and 5.6\,$\pm$\,1.9\%, respectively; mean\,$\pm$\,std over seeds).

\subsection{Published Results}
\label{app:published}
\begin{table*}[tp]
\centering\footnotesize
\setlength{\tabcolsep}{4pt}
\renewcommand{\arraystretch}{1.12}
\begin{tabular}{@{}>{\raggedright\arraybackslash}p{0.25\textwidth}>{\raggedright\arraybackslash}p{0.25\textwidth}>{\raggedright\arraybackslash}p{0.45\textwidth}@{}}
\toprule
\rowcolor{tablehead}Cited system & Source & Evaluation protocol \\
\midrule
GPT-Realtime, Gemini Live 2.5 and 3.1, Grok, Ultravox, Cascaded & \citet{lin2026fdbv3}, Table~2 & GPT-4o judge. GPT-Realtime is gpt-realtime-1.5 and Ultravox v0.7; Cascaded is Whisper \citep{radford2023whisper}, GPT-4o and OpenAI TTS in sequence. \\
MOSS-Live & \citet{mosi2026mosslive}, developer report, accessed 12 November 2026 & The developer report gives a tool-call success rate of 51.0\%. \\
Gander, GPT-5.6 as the back brain & \citet{orantqing2026gander}, Table~3 & The benchmark's scoring scripts with GPT-4o as the judge. \\
NemotronLabs VoiceChat & \citet{nemotronlabs2026voicechat}, Table~4 & Tool-selection F1, argument accuracy and Pass@1. \\
RV-Audio, offline & \citet{venus2026realtimevenus}, Table~7 (arXiv v3) & Offline inference with the tool definitions in the prompt; judge not named. We evaluate RV-Audio + Harness in our streaming pipeline separately (Appendix~\ref{app:eval}). \\
\citeauthor{hu2026frontendbackend}, Qwen3-235B-A22B backend & \citet{hu2026frontendbackend}, Table~2 (arXiv v2) & Tool-selection accuracy, not F1; GPT-4o judges the agent's text, transcribed by an external ASR system. \\
DuplexAgent-RSI on VoiceChat and on Qwen-Realtime; the Realtime-Venus system on Realtime-Omni-9B & \citet{shen2026duplexagentrsi}, Table~4 & Arguments are compared by exact string matching; Pass@1 is 47.0\% and 49.0\% for DuplexAgent-RSI and 1.0\% for the Realtime-Venus system. \\
MiniCPM-o~4.5, FDB v1.0 & \citet{adaptduplex2026}, Table~1 &  The three takeover rates compared in Table~\ref{tab:calibration}. \\
\bottomrule
\end{tabular}
\caption{Sources and evaluation protocols for published results. FDB v3 results appear in Table~\ref{tab:main}; MiniCPM-o~4.5 provides the FDB v1.0 reference in Table~\ref{tab:calibration}. Each result uses its source's evaluation setup and scoring protocol.}
\label{tab:cited}
\end{table*}

Table~\ref{tab:cited} records the sources and evaluation settings of the published results in Table~\ref{tab:main}.

\subsection{Turn-Taking and General Ability}

\begin{table}[tbp]
\centering
\footnotesize
\setlength{\tabcolsep}{6pt}
\renewcommand{\arraystretch}{1.12}
\begin{tabular}{@{}lcc@{}}
\toprule
\rowcolor{tablehead}
FDB v1.0 metric & Reported & Ours \\
\midrule
Turn-taking TOR$\uparrow$
  & 88.2 & 86.3\,$\pm$\,4.0 \\
\midrule[0.25pt]
Pause TOR, real$\downarrow$
  & 46.3 & 38.7\,$\pm$\,1.8 \\
\midrule[0.25pt]
Interruption TOR$\uparrow$
  & 98.5 & 99.2\,$\pm$\,1.4 \\
\bottomrule
\end{tabular}

\caption{
FDB v1.0 reproduction check on MiniCPM-o~4.5
for the three takeover rates used in our analysis,
compared with \citet[Table~1]{adaptduplex2026}.
Ours: mean\,$\pm$\,std over three seeds;
rates in \%.
All three differ from the reported values by
less than 10 percentage points.
TOR: takeover rate;
$\uparrow$/$\downarrow$: higher/lower is better.
}
\label{tab:calibration}
\end{table}

We evaluate FDB v1.0 \citep{lin2025fullduplexbench} with its official scripts and Parakeet transcripts \citep{rekesh2023fastconformer}, using 727 inputs per seed and three seeds.
Output audio is recorded on a clock aligned with the input audio, as required by the pause and turn-boundary metrics.
We check the pipeline on MiniCPM-o~4.5 \citep{cui2026minicpmo45} against \citet[Table~1]{adaptduplex2026}.

MiniCPM-o~4.5, RV-Audio and the three Duopra checkpoints run through the same streaming pipeline.
We also evaluate these five systems on FDB v1.5 \citep{lin2025fdbv15} (498 scenes and 996 recordings per seed) and VoiceBench \citep{chen2024voicebench} (7,783 items per seed), using three seeds per system and Qwen3.8-27B for judge-based metrics.
These evaluations compare conversational behavior and general ability across our training stages.
Most of RV-Audio's multiple-choice responses on VoiceBench are unparsed by the scorer; its reported score includes these parsing failures.

\section{Additional Results}
\label{app:results}

\subsection{Initiation vs.\ Execution (RQ2)}

\begin{figure*}[t]
\centering
\figbox{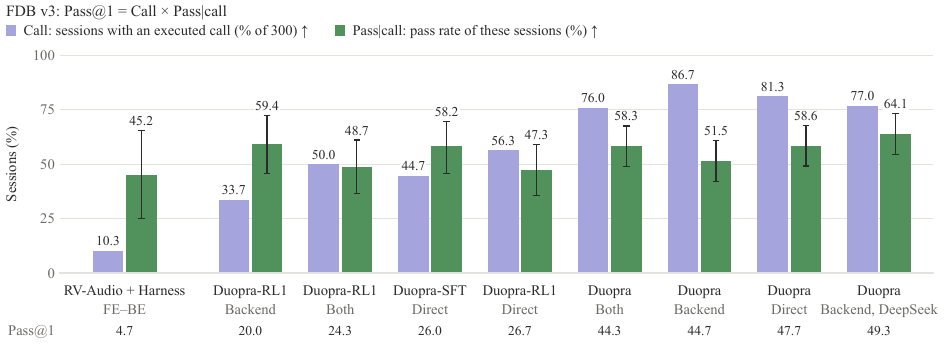}{\textwidth}{2.3in}
\caption{FDB v3 Pass@1 decomposed into Call (sessions with an executed domain-tool call) and Pass$|$call (success conditional on such a call; 95\% recording-level bootstrap intervals). The configurations are ordered by Pass@1 and use 300 sessions each (100 recordings, three decoding seeds). Low Call constrains success even when Pass$|$call is moderate. FE--BE: RV-Audio's request interface; backend: Qwen3-8B unless DeepSeek is named.}
\label{fig:decomp}
\end{figure*}

\paragraph{Coverage vs.\ conditional success.}
Figure~\ref{fig:decomp} shows why Pass@1 alone can conceal the failure source: a frontend may rarely reach an executed call, whereas another may call frequently but still fail conditional on calling. The identity Pass@1 = Call $\times$ Pass$|$call holds for every configuration; the latter factor is conditioned on a different called subset in each row.

\paragraph{Matched sessions locate the gain.}
For RV-Audio~+ Harness, the 31 sessions with a domain call yield 14 baseline passes and 12 Duopra passes; Duopra's net benefit instead comes from the other 269 sessions, where it succeeds on 122 requests that the baseline never executes. For Duopra-RL1 (Backend), Duopra passes 56 versus 60 on the 101 baseline-called sessions and 78 of the remaining 199 baseline-no-call sessions.

\paragraph{Hand-offs vs.\ executed calls.}
The RV-Audio~+ Harness frontend sends 34 hand-offs across 33 sessions, but domain tools run in only 31. Accordingly, its hand-off ceiling is based on 33 sessions, while its Call uses 31 (Table~\ref{tab:main}; Figure~\ref{fig:decomp}). With DeepSeek, Duopra's conditional success rises while its call coverage falls.

\begin{table*}[tp]
\centering\footnotesize
\setlength{\tabcolsep}{2.3pt}
\begin{tabular}{@{}lllcccccccccc@{}}
\toprule
\rowcolor{tablehead}& & & & \multicolumn{4}{c}{Outcome (sessions)} & \multicolumn{2}{c}{No call} & \multicolumn{3}{c}{Passes by calls$\!\uparrow$}\\
\cmidrule(lr){5-8}\cmidrule(lr){9-10}\cmidrule(l){11-13}
\rowcolor{tablehead}Frontend & Exposure & Backend LLM & Pass@1$\!\uparrow$ & Pass$\!\uparrow$ & Wrong arg. & Missing & Extra & Spoke & Silent & 1 & 2 & 3\\
\midrule
RV-Audio + Harness & FE--BE  & Qwen3-8B & 4.7\,$\pm$\,1.2  & 14  & 9  & 7  & 1  & 144 & 125 & 11  & 1  & 2\\
Duopra-SFT         & Direct  & none     & 26.0\,$\pm$\,3.6 & 78  & 17 & 38 & 1  & 118 & 48  & 76  & 2  & 0\\
\dmid              & Direct  & none     & 26.7\,$\pm$\,0.6 & 80  & 17 & 72 & 0  & 85  & 46  & 80  & 0  & 0\\
\dmid              & Both    & Qwen3-8B & 24.3\,$\pm$\,1.5 & 73  & 17 & 60 & 0  & 94  & 56  & 73  & 0  & 0\\
\dmid              & Backend & Qwen3-8B & 20.0\,$\pm$\,1.0 & 60  & 27 & 5  & 9  & 119 & 80  & 48  & 11 & 1\\
\rowcolor{oursfill}Duopra             & Direct  & none     & 47.7\,$\pm$\,3.5 & 143 & 58 & 40 & 3  & 39  & 17  & \textbf{115} & 13 & \textbf{15}\\
\rowcolor{oursfill}Duopra             & Backend & Qwen3-8B & 44.7\,$\pm$\,5.7 & 134 & 85 & 25 & 16 & \multicolumn{2}{c}{40} & 114 & 11 & 9\\
\rowcolor{oursfill}Duopra             & Both    & Qwen3-8B & 44.3\,$\pm$\,0.6 & 133 & 53 & 39 & 3  & \multicolumn{2}{c}{72} & 105 & 16 & 12\\
\rowcolor{oursfill}Duopra             & Backend & DeepSeek & \textbf{49.3}\,$\pm$\,3.5 & \textbf{148} & 46 & 27 & 10 & \multicolumn{2}{c}{69} & \textbf{115} & \textbf{19} & 14\\
\rowcolor{oursfill}Duopra             & Both    & DeepSeek & 44.7\,$\pm$\,2.5 & 134 & 52 & 37 & 5  & \multicolumn{2}{c}{72} & 110 & 10 & 14\\
\bottomrule
\end{tabular}
\caption{FDB v3 session outcomes (300 sessions per configuration). Pass@1: decoding-seed mean$\pm$std. Outcome classes are exclusive (Appendix~\ref{app:eval}); merged no-call cells include sessions with and without speech. Successes by required call count use 198 one-call, 54 two-call and 48 three-call sessions. FE--BE: RV-Audio's original request interface. No-call and wrong-argument outcomes reveal distinct points of failure.}
\label{tab:fdbv3-runs}
\end{table*}

\paragraph{Tool visibility and routing.}
Table~\ref{tab:fdbv3-runs} reports the complete session outcomes. Under Both, Duopra delegates in one Qwen3-8B session and none with DeepSeek. Of its 133 successes with Qwen3-8B, at least 132 therefore occur without delegation; at least 27 of the 28 multi-call successes do so. All 134 DeepSeek-Both successes occur without delegation. Duopra-RL1 also rarely delegates under Both but succeeds less often; therefore, low delegation frequency by itself does not establish routing competence. Under Backend, restricting the frontend to \tool{query} tests whether its requests suffice for backend execution rather than its ability to choose between routes.

\paragraph{Latency and argument fidelity.}
\label{app:latency}
Backend has higher factual-answer latency than Both with either backend LLM, and for Qwen3-8B it also has a higher wrong-argument share. The observed latency is measured until the factual answer, including runtime and tool execution (Appendix~\ref{app:eval}). The respective latency values are 14.85$\pm$0.53 vs.\ 10.42$\pm$0.33\,s for Qwen3-8B, and 13.60$\pm$0.60 vs.\ 9.95$\pm$0.93\,s for DeepSeek (Backend vs.\ Both, decoding-seed mean$\pm$std). The wrong-argument comparisons are given in Table~\ref{tab:ablation}b; they describe the tested exposures rather than a controlled cost of delegation.

\subsection{Training Stages and Failure Modes (RQ3)}
\begin{table*}[tp]
\centering\footnotesize
\setlength{\tabcolsep}{3.2pt}
\begin{tabular}{@{}lccccccccc@{\hspace{10pt}}ccc@{}}
\toprule
\rowcolor{tablehead}& \multicolumn{9}{c}{FDB v3} & \multicolumn{3}{c}{FDB v1.0 TOR}\\
\cmidrule(lr){2-10}\cmidrule(l){11-13}
\rowcolor{tablehead}(a) Model & Pass@1$\!\uparrow$ & Easy$\!\uparrow$ & Med.$\!\uparrow$ & Hard$\!\uparrow$ & 2--3 calls$\!\uparrow$ & Tool$\!\uparrow$ & Arg.$\!\uparrow$ & Resp.$\!\uparrow$ & Take$\!\uparrow$ & Turn$\!\uparrow$ & Pause$\!\downarrow$ & Intr.$\!\uparrow$\\
\midrule
RV-Audio   & 0.0\,$\pm$\,0.0  & 0.0  & 0.0  & 0.0  & 0.0  & 0.0  & 0.0  & 17.0 & 51.3 & 59.1 & 18.1 & 79.5\\
Duopra-SFT & 26.0\,$\pm$\,3.6 & 43.5 & 23.5 & 7.8  & 2.0  & 39.2 & 30.1 & 56.8 & 82.7 & 75.1 & 9.3  & \textbf{97.0}\\
\dmid      & 26.7\,$\pm$\,0.6 & 47.2 & 25.5 & 3.3  & 0.0  & 44.7 & 32.2 & 59.1 & 82.0 & \textbf{75.6} & 5.6 & 96.5\\
\rowcolor{oursfill}Duopra & \textbf{47.7}\,$\pm$\,3.5 & \textbf{59.3} & \textbf{52.0} & \textbf{28.9} & \textbf{27.5} & \textbf{76.7} & \textbf{56.2} & \textbf{66.3} & \textbf{93.0} & \textbf{75.6} & \textbf{5.4} & 91.3\\
\midrule
\multicolumn{13}{@{}l}{\textit{Reference: frontend--backend agents with the Qwen3-8B backend}}\\
RV-Audio + Harness & 4.7\,$\pm$\,1.2 & 7.4 & 3.9 & 2.2 & 2.9 & 9.5 & 5.9 & 22.9 & 50.0 & \na & \na & \na\\
DuplexOmni & 17.0\,$\pm$\,2.0 & 28.7 & 14.7 & 5.6 & \na & 60.0 & 27.5 & 42.0 & \na & \na & \na & \na\\
\bottomrule
\end{tabular}

\vspace{7pt}

\begin{minipage}[t]{0.49\textwidth}
\vspace{0pt}\centering
\begin{tabular}{@{}lr@{}}
\toprule
\rowcolor{tablehead}(b) First $-$ second & $\Delta$ [95\% CI]\\
\midrule
\multicolumn{2}{@{}l}{\textit{Frontend, domain tools behind Qwen3-8B (Backend)}}\\
Duopra $-$ RV-Audio + Harness           & +40.0 [31.3, 48.7]\\
Duopra $-$ \dmid{}                      & +24.7 [17.3, 32.3]\\
\dmid{} $-$ RV-Audio + Harness             & +15.3 [9.0, 22.0]\\
\multicolumn{2}{@{}l}{\textit{Backend LLM, Duopra (Backend)}}\\
DeepSeek $-$ Qwen3-8B                      & +4.7 [$-$2.3, 11.7]\\
\multicolumn{2}{@{}l}{\textit{Tool exposure, \dmid{} (Qwen3-8B)}}\\
Direct $-$ Backend                         & +6.7 [0.3, 13.0]\\
Direct $-$ Both                            & +2.3 [$-$1.7, 6.0]\\
\multicolumn{2}{@{}l}{\textit{Tool exposure, Duopra}}\\
Backend $-$ Both (Qwen3-8B)                & +0.3 [$-$7.3, 8.0]\\
Direct $-$ Both (Qwen3-8B)                 & +3.3 [$-$1.0, 7.7]\\
Direct $-$ Both (DeepSeek)                 & +3.0 [$-$1.3, 7.3]\\
\bottomrule
\end{tabular}
\end{minipage}\hfill
\begin{minipage}[t]{0.49\textwidth}
\vspace{0pt}\centering
\begin{tabular}{@{}lr@{}}
\toprule
\rowcolor{tablehead}(b, cont.) First $-$ second & $\Delta$ [95\% CI]\\
\midrule
\multicolumn{2}{@{}l}{\textit{Training stage (Direct)}}\\
\dmid{} $-$ Duopra-SFT (Stage 2a)          & +0.7 [$-$5.0, 6.0]\\
Duopra $-$ \dmid{} (Stage 2b)              & +21.0 [15.3, 27.0]\\
Duopra $-$ Duopra-SFT                      & +21.7 [14.0, 29.3]\\
\quad recordings needing one call          & +19.7 [10.1, 29.3]\\
\quad recordings needing 2--3 calls        & +25.5 [13.7, 38.2]\\
\multicolumn{2}{@{}l}{\textit{Wrong argument, Duopra: Backend $-$ Both}}\\
Qwen3-8B                                   & +10.7 [4.3, 17.3]\\
DeepSeek                                   & $-$2.0 [$-$8.3, 4.3]\\
\multicolumn{2}{@{}l}{\textit{Pass$|$call, Duopra (Backend)}}\\
DeepSeek $-$ Qwen3-8B                      & +12.5 [4.8, 20.4]\\
\bottomrule
\end{tabular}
\end{minipage}
\caption{Training-checkpoint comparisons. (a)~FDB v3 Direct scores and three calibrated FDB v1.0 takeover rates (mean$\pm$std over decoding seeds); 2--3 calls: 102 multi-call sessions. (b)~Paired differences with recording-bootstrap 95\% intervals. Wrong argument is the share of all sessions with correct tools but incorrect arguments; Pass$|$call conditions on an executed domain call. Configurations: Table~\ref{tab:model-configurations}; metric definitions: Appendix~\ref{app:eval}. \na: not evaluated.}
\label{tab:ablation}
\end{table*}

\paragraph{Invocation vs.\ completion.}
Table~\ref{tab:ablation} shows that the intermediate Duopra-RL1 checkpoint increases the frequency of calls but scarcely changes task success relative to Duopra-SFT. Duopra improves Pass@1 most on medium, hard, and multi-call requests. As Stage~2b also introduces new tasks and reward rules, this comparison attributes the observed change to the overall training stage, not to a single reward term.

\paragraph{Shifting error classes.}
In Figure~\ref{fig:analysis}, no-call sessions decline markedly across checkpoints, and requests requiring several calls become solvable more often. However, incorrect arguments remain common once the correct tool set is reached. The figure and Table~\ref{tab:fdbv3-runs} separate those error classes rather than treating every unsuccessful session as a failure to act.

\subsection{Turn-Taking and General Ability (RQ4)}
\begin{table*}[tp]
\centering\footnotesize
\setlength{\tabcolsep}{5pt}
\begin{tabular}{@{}lcccc>{\columncolor{oursfill}}c@{}}
\toprule
\rowcolor{tablehead}Metric & MiniCPM-o 4.5 & RV-Audio & Duopra-SFT & \dmid & Duopra\\
\midrule
\rowcolor{tablealt}\multicolumn{6}{@{}l}{\textit{(a) FDB v1.0}}\\
Turn-taking TOR$^{c}\!\uparrow$            & \textbf{86.3}\,$\pm$\,4.0 & 59.1\,$\pm$\,2.6 & 75.1\,$\pm$\,0.5 & 75.6\,$\pm$\,0.8 & 75.6\,$\pm$\,3.0\\
Pause TOR, real$^{c}\!\downarrow$          & 38.7\,$\pm$\,1.8 & 18.1\,$\pm$\,2.3 & 9.3\,$\pm$\,0.9 & 5.6\,$\pm$\,2.1 & \textbf{5.4}\,$\pm$\,0.3\\
Interruption TOR$^{c}\!\uparrow$           & \textbf{99.2}\,$\pm$\,1.4 & 79.5\,$\pm$\,3.0 & 97.0\,$\pm$\,1.5 & 96.5\,$\pm$\,0.9 & 91.3\,$\pm$\,0.6\\
Pause TOR, synthetic$\downarrow$           & 10.5\,$\pm$\,1.1 & 10.5\,$\pm$\,1.8 & 5.4\,$\pm$\,1.1 & 3.2\,$\pm$\,1.1 & 2.4\,$\pm$\,1.1\\
Turn-taking latency (s)$\downarrow$        & 2.23\,$\pm$\,0.08 & 2.27\,$\pm$\,0.15 & 2.64\,$\pm$\,0.04 & 2.75\,$\pm$\,0.07 & 2.75\,$\pm$\,0.05\\
Interruption latency (s)$\downarrow$       & 1.92\,$\pm$\,0.09 & 2.58\,$\pm$\,0.05 & 1.99\,$\pm$\,0.05 & 2.18\,$\pm$\,0.02 & 2.28\,$\pm$\,0.03\\
Interruption quality$^{\ddagger}$ (0--5)$\uparrow$ & 4.14\,$\pm$\,0.05 & 3.65\,$\pm$\,0.04 & 4.30\,$\pm$\,0.13 & 4.21\,$\pm$\,0.06 & 4.38\,$\pm$\,0.13\\
Backchannel TOR$\downarrow$                & 58.8\,$\pm$\,6.4 & 60.0\,$\pm$\,4.8 & 0.0$^{b}$ & 0.0$^{b}$ & 0.0$^{b}$\\
Backchannel frequency (/s)                 & 0.063\,$\pm$\,0.006 & 0.106\,$\pm$\,0.030 & 0$^{b}$ & 0$^{b}$ & 0$^{b}$\\
Backchannel JSD$\downarrow$                & 0.794\,$\pm$\,0.014 & 0.789\,$\pm$\,0.032 & 1.000$^{b}$ & 1.000$^{b}$ & 1.000$^{b}$\\
\midrule
\rowcolor{tablealt}\multicolumn{6}{@{}l}{\textit{(b) FDB v1.5$^{\ddagger}$}}\\
Response after an interruption$\uparrow$   & 47.5\,$\pm$\,2.6 & 35.2\,$\pm$\,2.1 & 69.5\,$\pm$\,2.3 & 67.0\,$\pm$\,2.3 & 66.0\,$\pm$\,1.0\\
Resumption after a backchannel$\uparrow$   & 87.8\,$\pm$\,5.7 & 36.1\,$\pm$\,5.2 & 95.9\,$\pm$\,2.7 & 96.6\,$\pm$\,0.6 & 96.3\,$\pm$\,0.6\\
Resumption, user talks to others$\uparrow$ & 56.7\,$\pm$\,2.1 & 43.3\,$\pm$\,8.6 & 45.3\,$\pm$\,2.3 & 56.3\,$\pm$\,3.2 & 52.0\,$\pm$\,3.5\\
Resumption, background speech$\uparrow$    & 59.0\,$\pm$\,3.6 & 43.7\,$\pm$\,1.5 & 68.3\,$\pm$\,3.2 & 70.7\,$\pm$\,2.1 & 71.0\,$\pm$\,2.0\\
\midrule
\rowcolor{tablealt}\multicolumn{6}{@{}l}{\textit{(c) VoiceBench$^{\ddagger}$ (all $\uparrow$)}}\\
Mean of nine categories                    & 64.9\,$\pm$\,0.2 & 37.9\,$\pm$\,0.4 & 52.5\,$\pm$\,0.1 & 52.9\,$\pm$\,0.2 & 55.9\,$\pm$\,0.2\\
MMSU                                       & 57.4\,$\pm$\,0.2 & 26.6\,$\pm$\,0.7 & 54.7\,$\pm$\,0.4 & 56.4\,$\pm$\,0.6 & 62.5\,$\pm$\,0.1\\
OpenBookQA                                 & 73.1\,$\pm$\,1.3 & 24.6\,$\pm$\,1.4 & 69.2\,$\pm$\,1.5 & 70.8\,$\pm$\,1.0 & 77.4\,$\pm$\,1.3\\
BBH                                        & 63.7\,$\pm$\,1.7 & 50.6\,$\pm$\,1.9 & 59.7\,$\pm$\,0.8 & 61.0\,$\pm$\,1.0 & 60.0\,$\pm$\,0.4\\
IFEval                                     & 55.0\,$\pm$\,2.0 & 24.9\,$\pm$\,1.4 & 33.2\,$\pm$\,0.9 & 34.1\,$\pm$\,1.6 & 39.2\,$\pm$\,0.9\\
SD-QA                                      & 48.5\,$\pm$\,1.7 & 31.5\,$\pm$\,1.9 & 27.0\,$\pm$\,0.9 & 24.4\,$\pm$\,1.3 & 27.7\,$\pm$\,1.3\\
AlpacaEval                                 & 72.4\,$\pm$\,1.1 & 41.6\,$\pm$\,1.1 & 54.6\,$\pm$\,0.1 & 54.9\,$\pm$\,0.2 & 57.7\,$\pm$\,0.3\\
CommonEval                                 & 64.3\,$\pm$\,0.3 & 41.0\,$\pm$\,0.8 & 45.0\,$\pm$\,0.5 & 44.3\,$\pm$\,1.1 & 49.5\,$\pm$\,0.4\\
WildVoice                                  & 58.3\,$\pm$\,0.2 & 34.5\,$\pm$\,0.5 & 41.0\,$\pm$\,0.4 & 40.0\,$\pm$\,0.4 & 42.2\,$\pm$\,0.4\\
AdvBench (refusal rate)                    & 91.2\,$\pm$\,1.3 & 65.6\,$\pm$\,1.3 & 88.4\,$\pm$\,1.2 & 90.6\,$\pm$\,1.0 & 87.1\,$\pm$\,1.8\\
\bottomrule
\end{tabular}
\caption{Conversational behavior and general ability in our streaming pipeline (three seeds, mean\,$\pm$\,std; rates and scores in \% unless stated otherwise). TOR: takeover rate; RV-Audio: Realtime-Venus-Audio. $^{c}$Takeover rates used in the main-text comparison, calibrated on MiniCPM-o~4.5 (Table~\ref{tab:calibration}); bold marks the best value on each calibrated row. $^{\ddagger}$Judged wholly or partly by Qwen3.8-27B. $^{b}$No backchannels are produced by any of the three post-trained
checkpoints on the 55 FDB v1.0 backchannel items; the reported frequency
and TOR are zero, and JSD is 1.000.
These rows show a selective cost of task training and measure a different behavior from resumption
after a user's backchannel in FDB v1.5. Model configurations and VoiceBench parsing limitations appear in Appendix~\ref{app:eval}. $\uparrow$/$\downarrow$: higher/lower is better.}
\label{tab:retention}
\end{table*}

\paragraph{Behavior-specific retention.}
Table~\ref{tab:retention} shows improvements in several turn-taking and overlap metrics in our pipeline, but not uniform retention of conversational ability. All three post-trained checkpoints produce no backchannels on the tested FDB v1.0 items; FDB v1.5 instead measures whether speech continues after a \emph{user's} backchannel, which is a distinct behavior. Interruption response also declines from Duopra-RL1 to Duopra. VoiceBench results provide a separate check on broader spoken-answer ability, with parsing limitations for the base model noted in Appendix~\ref{app:eval}.

\subsection{Talker Conditioning and Repair (RQ4)}
\label{app:projector}
At spoken-token position $j$, talker conditioning is
$c_j=\mathrm{emb}_{\mathrm{text}}(\mathrm{token}_j)+\mathrm{L2norm}(P_\phi(h_j))$,
where the thinker states and projector parameters are taken from the evaluated configuration. We test both the change in this signal after thinker training and the effect of repairing only its projector (Section~\ref{sec:projector}).

\begin{figure*}[t]
\centering
\figbox{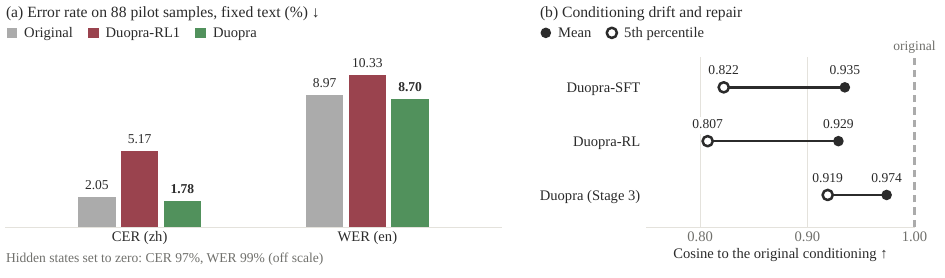}{\textwidth}{1.85in}
\caption{Talker conditioning and speech errors across training stages. (a)~CER and WER on 88 fixed-text samples for RV-Audio
(\textit{Original}), Duopra-RL1 with the base projector, and Duopra
with the Stage-3 projector.
(b)~Mean and 5th-percentile cosine similarity to the talker conditioning
of RV-Audio on 2,000 session replays with
identical prefixes. Duopra-RL uses the base projector, and Duopra uses
the Stage-3 projector with the same Duopra-RL thinker.
The dashed line marks cosine similarity of~1.}
\label{fig:projector}
\end{figure*}

\paragraph{Conditioning drift.}
Figure~\ref{fig:projector}b compares the thinker checkpoints using the base projector. Their mean cosines to the talker conditioning of RV-Audio (\textit{original} in the figure) are 0.935 (Duopra-SFT) and 0.929 (Duopra-RL), whereas the Stage-3 projector brings that cosine to 0.974 on the same replay analysis. The 88 fixed-text speech samples in panel (a) compare different thinker--projector pairs and therefore cannot by themselves isolate the projector's causal effect.

\paragraph{Repair with a fixed thinker.}
For the Stage-2b thinker, mean cosine on the feature validation set rises from 0.930 to 0.973 after repair; this set is distinct from the 2,000-session replay analysis above. We select the final projector from four candidates using 221 fixed-trajectory utterances, and report pooled CER/WER on that \emph{selection set} in Table~\ref{tab:stage3-speech}. Text, control tokens, unit boundaries, speaker and synthesis settings are fixed in this comparison; the base-projector row is the ablation and the Stage-3 row is Duopra.

\begin{table}[tbp]
\centering\footnotesize
\setlength{\tabcolsep}{4.5pt}
\begin{tabular}{lcccc}
\toprule
\rowcolor{tablehead} & \multicolumn{2}{c}{Error rate (\%)} & \multicolumn{2}{c}{ECAPA cosine}\\
\cmidrule(lr){2-3}\cmidrule(lr){4-5}
\rowcolor{tablehead}Projector & CER$\downarrow$ & WER$\downarrow$ & zh$\uparrow$ & en$\uparrow$\\
\midrule
Base projector & 6.01 & 9.74 & 0.778 & 0.636\\
\rowcolor{oursfill}Stage-3 projector & \textbf{2.58} & \textbf{7.66} & \textbf{0.782} & \textbf{0.648}\\
\bottomrule
\end{tabular}
\caption{Speech repair on the 221-utterance projector-\emph{selection} set (122 Chinese / 6,156 characters; 99 English / 2,833 words). Both rows use the same Duopra-RL thinker, fixed text and synthesis schedule; only its projector differs. ECAPA cosine: similarity to the native voice.}
\label{tab:stage3-speech}
\end{table}

\paragraph{Selection-set intelligibility.}
Table~\ref{tab:stage3-speech} shows lower Chinese character and English word error rates with the repaired projector, alongside small ECAPA cosine changes. Because the same utterances selected the candidate, the reported reduction should be interpreted as selection-set evidence rather than an independent generalization estimate.

\subsection{Case Examples}
\label{app:cases}
\begin{table*}[tp]
\centering
\footnotesize
\setlength{\tabcolsep}{4pt}
\renewcommand{\arraystretch}{1.14}

\newcommand{\caserule}{%
  \arrayrulecolor{black!25}%
  \specialrule{0.3pt}{4pt}{4pt}%
  \arrayrulecolor{black}%
}

\begin{tabularx}{\textwidth}{@{}
  >{\raggedright\arraybackslash}p{0.37\textwidth}
  >{\raggedright\arraybackslash}p{0.29\textwidth}
  >{\raggedright\arraybackslash}X
@{}}
\toprule
\rowcolor{tablehead}
User request & Executed calls & Agent's speech \\
\midrule

\textbf{(a) Backend; pass}\par\smallskip
\textit{Um... so you know... search for a laptop --- wait... no,... um...... um, I actually need a tablet, not a laptop. And add 2 of them to the cart. Actually, just add 1 instead, I only have budget for one right now.}
&
\tool{search\_products}\par
(query: tablet)\par\smallskip
\tool{add\_to\_cart}\par
(product\_id: tablet\_001, quantity: 1)\par\smallskip
Both at 30.5\,s.
&
I will search for a tablet and add one to your cart. The tablet tablet premium has been successfully added to your cart. The total price is \$99. \\

\caserule

\textbf{(b) Both; pass}\par\smallskip
\textit{Um... so hmm... hmm... well... could you track order B-O-B for me first? Then... um,... um, I also want to search for a nice watch under 200 dollars. And once you find something good, uh, go ahead and add it to my cart.}
&
\tool{track\_order}\par
(order\_id: BOB) at 36.0\,s\par\smallskip
\tool{search\_products}\par
(query: watch, max\_price: 200) at 43.2\,s\par\smallskip
\tool{add\_to\_cart}\par
(product\_id: PROD1, quantity: 1) at 51.0\,s
&
I will track the order and search for a watch. Your order B B is out for delivery. I found a watch named Watch Premium for \$190. The watch has been added to your cart. \\

\caserule

\textbf{(c) Backend; wrong argument}\par\smallskip
\textit{Hmm... um... so like... first, search for a gaming mouse ---... um, I want to see what you have. Then add the cheapest option to my cart so I can buy it later. And while you're at it, could you also track my order P-O-9-9-9? I've been waiting on that one.}
&
\tool{search\_products}\par
(query: gaming mice) at 29.0\,s\par\smallskip
\tool{add\_to\_cart}\par
(product\_id: PROD1, quantity: 1) at 30.3\,s\par\smallskip
\tool{track\_order}\par
(order\_id: P0999) at 31.4\,s
&
I will search for gaming mice. Add the cheapest one to your cart and check the status of order P0999. The order P09999 is currently out for delivery. The Gaming Mouse Gaming Mice Premium has been added to the cart with a total cost. \\

\caserule

\textbf{(d) Backend; no call}\par\smallskip
\textit{Uh... like, you know... um... so search for a coffee maker first. If you find one that's under 50 dollars, go ahead and add 2... um, to the cart because they'd make great gifts. But if everything's over 50, just forget the coffee maker and track order W-W-W for me instead.}
&
\textit{none}
&
I will search for a coffee maker under fifty dollars and add two to the cart. If everything is over fifty, I will forget the coffee maker and track order www for me instead. \\

\bottomrule
\end{tabularx}
\caption{Selected hard FDB v3 sessions from Duopra with Qwen3-8B as the backend (seed 641; case (d): seed 643). User requests are evaluation-record transcripts; agent speech is transcribed by ASR and retains recognition errors. Call times are measured from session onset. In (c), the executed order ID substitutes zero for the user's letter O.}
\label{tab:cases}
\end{table*}

Table~\ref{tab:cases} illustrates both successful routes and two failure modes on hard FDB v3 requests. The request in case (a) matches the public \tool{ecommerce\_19} scenario used schematically in Figure~\ref{fig:timeline}. The backend can complete a multi-step request from a faithful delegation (a), while the frontend can execute a sequence of direct calls (b). The failures involve an alphanumeric argument substitution (c) and a spoken plan without any executed call (d). 

\section{Additional Related Work}
\label{app:related}

\paragraph{Tool use and optimization.}
Text-based tool use spans self-supervised API learning \citep{schick2023toolformer}, interleaved reasoning and acting \citep{yao2023react}, and instruction tuning \citep{qin2024toolllm}.
BFCL \citep{patil2025bfcl} and $\tau$-bench \citep{yao2024taubench} evaluate function calls and interactive task completion.
GRPO retains PPO's clipped objective \citep{schulman2017ppo} while replacing its value model with group-relative advantages \citep{shao2024deepseekmath}.
Dr.~GRPO removes response-length and reward-standard-deviation normalizations \citep{liu2025drgrpo}; DAPO uses dynamic sampling to exclude groups with identical rewards \citep{yu2026dapo}.
Duopra applies this optimization framework to streaming sessions, with rewards for timely action, grounded arguments and task completion.

\paragraph{Spoken agents and evaluation.}
Alongside FDB v3 \citep{lin2026fdbv3}, benchmarks assess spoken tool calls \citep{voiceagentbench2025,audio2tool2026} and end-to-end voice-agent performance \citep{ray2026tauvoice,bogavelli2026evabench}.
Related systems delegate reasoning and tool use to background models \citep{thinkingmachines2026interactionmodels,orantqing2026gander} or retrieve knowledge asynchronously \citep{chien2026moshirag}.
Our analysis separates frontend action initiation from subsequent tool execution using the resulting call traces.

\paragraph{Sufficiency and dialogue state.}
Information sufficiency connects our setting to slot filling in schema-guided dialogue \citep{rastogi2020sgd} and missing-information tests for tool use \citep{patil2025bfcl,lu2025toolsandbox}.
Our formulation (Section~\ref{sec:problem}) tracks the latest spoken values and corrections; delegated requests must also carry required dependencies. The same sufficiency rule applies to both routes.

\end{document}